\documentclass[lettersize,journal]{IEEEtran}
\usepackage{amsmath,amsfonts}
\usepackage{algorithmic}
\usepackage{algorithm}
\usepackage{array}
\usepackage[caption=false,font=normalsize,labelfont=sf,textfont=sf]{subfig}
\usepackage{textcomp}
\usepackage{stfloats}
\usepackage{url}
\usepackage{verbatim}
\usepackage{graphicx}
\usepackage{cite}
\usepackage{colortbl}
\usepackage{booktabs}
\usepackage{multirow}
\usepackage{CJKutf8}

\usepackage[inkscapearea=page]{svg}

\newcommand{\lpipe}{\rule[-0.3ex]{0.3pt}{2ex}}

\begin{document}

\title{CReMa: Crisis Response through Computational Identification and Matching of Cross-Lingual Requests and Offers Shared on Social Media}


\author{Rabindra Lamsal, Maria Rodriguez Read, Shanika Karunasekera, Muhammad Imran
\thanks{This work was supported by the Melbourne Research Scholarship from the University of Melbourne. (\textit{Corresponding author: Rabindra Lamsal})}%

\thanks{Rabindra Lamsal, Maria Rodriguez Read, Shanika Karunasekera are with the School of Computing and Information Systems, The University of Melbourne, Carlton, VIC 3052, Australia (e-mail: r.lamsal@unimelb.edu.au, maria.read@unimelb.edu.au, karus@unimelb.edu).}%

\thanks{Muhammad Imran is with Qatar Computing Research Institute,  Hamad Bin Khalifa University, Qatar (email: mimran@hbku.edu.qa).}
}

\markboth{Accepted to IEEE Transactions on Computational Social Systems}%
{Shell \MakeLowercase{\textit{Lamsal et al.}}: Crisis Response through Computational Matching of Cross-Lingual Requests and Offers Shared on Social Media}


\maketitle

\begin{abstract}
During times of crisis, social media platforms play a crucial role in facilitating communication and coordinating resources. In the midst of chaos and uncertainty, communities often rely on these platforms to share urgent pleas for help, extend support, and organize relief efforts. However, the overwhelming volume of conversations during such periods can escalate to unprecedented levels, necessitating the automated identification and matching of requests and offers to streamline relief operations. Additionally, there is a notable absence of studies conducted in multi-lingual settings, despite the fact that any geographical area can have a diverse linguistic population. Therefore, we propose \textit{CReMa} (\underline{C}risis \underline{Re}sponse \underline{Ma}tcher), a systematic approach that integrates textual, temporal, and spatial features to address the challenges of effectively identifying and matching requests and offers on social media platforms during emergencies. Our approach utilizes a crisis-specific pre-trained model and a multi-lingual embedding space. We emulate human decision-making to compute temporal and spatial features and non-linearly weigh the textual features. The results from our experiments are promising, outperforming strong baselines. Additionally, we introduce a novel multi-lingual dataset simulating help-seeking and offering assistance on social media in 16 languages and conduct comprehensive cross-lingual experiments. Furthermore, we analyze a million-scale geotagged global dataset to understand patterns in seeking help and offering assistance on social media. Overall, these contributions advance the field of crisis informatics and provide benchmarks for future research in the area.
\end{abstract}

\begin{IEEEkeywords}
crisis embeddings, multi-lingual matching, CrisisTransformers, classification models, sentence encoders, vector search
\end{IEEEkeywords}

\section{Introduction}

Social media is an efficient medium for information sharing during crises; from natural disasters to public health emergencies, the platforms facilitate rapid communication and resource coordination influencing crisis response efforts worldwide \cite{imran2015processing,lamsal2022socially}. The importance of social media also lies in its ability to mobilize resources and connect individuals in need with those who can provide assistance. In the midst of chaos and uncertainty, communities often turn to social media platforms to share urgent pleas for help, offer support, and coordinate relief efforts. Whether it is a call for medical supplies, shelter, or assistance, these platforms serve as virtual hubs where individuals can broadcast their voices and leverage collective action for the greater good.

For effective emergency relief coordination through social media, the identification and matching of two crucial types of situational information are essential: requests and offers \cite{purohit2014emergency}. Requests indicate shortages of specific resources/services (such as shelter, clothing, food, or volunteers), while offers signify the availability or willingness to provide certain resources/services. These requests and offers may originate from individuals or organizations and can be made on behalf of oneself or others. For example, \texttt{Request}: \textit{Seeking assistance for food and essentials in the aftermath of the Hurricane.} \texttt{Offer}: \textit{If you have been affected by hurricane sandy...FEMA is giving \$200 food stamp vouchers at [address]}. Matching requests with related offers can greatly aid relief workers in delivering effective and efficient crisis response. However, the sheer volume of conversations during a crisis, which can reach hundreds of thousands or even millions \cite{qazi2020geocov19,lamsal2023billioncov}, requires the need for automated identification and matching of requests and offers to facilitate the coordination of relief operations. Several prior works have addressed the \textit{identification tasks} \cite{purohit2014emergency,nazer2016finding,ullah2021rweetminer}, while there have been limited studies attempting to address the \textit{matching tasks} \cite{purohit2014emergency,dutt2019utilizing}. This study focuses on both tasks and enhances prior research, establishing the work as a benchmark for future studies in the area.

To design text classifiers for the identification tasks, previous research has primarily relied on traditional machine learning classifiers or transformer-based models like BERT and RoBERTa, which are pre-trained on general text data. However, we utilize crisis-domain-specific pre-trained models developed in our earlier work (released as CrisisTransformers \cite{lamsal2024crisistransformers}), which were trained on over 15 billion tokens extracted from tweets related to 30+ crisis events. Furthermore, unlike previous studies that relied on traditional textual representation methods for the matching tasks, we employ a CrisisTransformers-based embedding model \cite{lamsal2023crosslingual}, whose embedding space has been restructured such that semantically related crisis-specific sentences are placed close together in the vector space. Another significant gap in the literature is the absence of studies conducted in a multi-lingual setting. Given that any geographical area can have a diverse linguistic population, and a social media platform can therefore receive requests and offers in different languages, processing only a particular language introduces a risk of overlooking critical information available in texts shared in other languages. To address this notable gap, our study utilizes a cross-lingual embedding space.

Overall, this study makes the following contributions to the existing \textbf{crisis informatics} literature:

\begin{enumerate}
\item We propose CReMa (\textbf{C}risis \textbf{Re}sponse \textbf{Ma}tcher), a systematic approach that integrates textual, temporal, and spatial features to identify and match requests and offers shared on social media during a crisis.

\item We manually curate a new multi-lingual dataset\footnote{Dataset will be made available for research purposes upon request.} (13k samples) simulating help-seeking and offering assistance scenarios on social media. Utilizing this dataset, we demonstrate that a cross-lingual embedding space can be effectively used to extract textual features for enhancing the matching task.

\item As a case study, we analyze a million-scale geotagged global dataset to explore the distribution of crisis communications regarding requests and offers on social media during the COVID-19 pandemic.
\end{enumerate}

The rest of the paper is organized as follows: Section \ref{related} discusses related work, Section \ref{method} discusses the method used by CReMa for identifying and matching requests and offers, Section \ref{result} presents results and discussions, and Section \ref{conclusion} concludes the study and provides future directions.

\section{Related work}
\label{related}
Multiple works have been done in the identification of help requests, offers, and matching respective help requests with potential offers. Reviewing the literature in the area, we first discuss the works that focused on the identification tasks. Subsequently, we discuss works related to matching tasks.

As an early work, Purohit et al. \cite{purohit2014emergency} curated and released sets of regular expressions and labeled datasets for identifying request and offer texts. They extracted n-grams as features and employed two binary classifiers in a cascade configuration, both based on Random Forest. The first classifier categorizes tweets into either ``request'' or ``non-exclusive request', and the second classifier takes ``non-exclusive request'' tweets, categorizing them as either ``offer'' or ``other'. Nazer et al. \cite{nazer2016finding} improved the help request classification performance by extending the features and training a Decision Tree classifier. They transitioned from using just n-grams to incorporating topics generated by a topic model, URLs, hashtags, mentions, posted time, retweet count, and author’s profile metadata. Furthermore, Ullah et al. \cite{ullah2021rweetminer} enhanced the classification by introducing rule-based features to n-grams and trained a Logistic Regression classifier. Each tweet considered for classification was checked against a set of regular expressions introduced by Purohit et al. \cite{purohit2014emergency}, and the frequency of matched expressions was used as a rule-based feature. Their proposed approach had a two-phase classification. In the first phase, an incoming tweet is checked to determine if it is a help request. If the tweet is identified as a help request, a classifier categorizes it into one of the ``resources'' categories defined in \cite{purohit2014emergency}. 

Dense vector representations have also been employed for the identification task. Devaraj et al. \cite{devaraj2020machine} demonstrated the efficacy of using word vectors from GloVe \cite{pennington2014glove} as features alongside n-grams and part-of-speech (POS) tags for categorizing tweets into ``urgent'' and ``not urgent'' categories. Similarly, He et al. \cite{he2017signals} utilized n-grams and word vectors from word2vec \cite{mikolov2013efficient} as features to train a binary XGBoost classifier for identifying whether a tweet contains ``logistical information'' (offering or requesting help as a single category). Transformer-based models have also been employed in designing classifiers for similar tasks. Zhou et al. \cite{zhou2022victimfinder} experimented with multiple pre-trained models, including BERT, RoBERTa, and XLNet, to binary classify tweets across four tasks, one of which is ``is the tweet asking for help?'. Their results show that the transformer-based models outperform the baselines Glove and ELMo \cite{peters2018deep}.

The studies \cite{purohit2014emergency, nazer2016finding, ullah2021rweetminer} mainly rely on sparse vector representations for training classifiers, which may limit their ability to capture nuanced semantic relationships. While some \cite{devaraj2020machine, he2017signals} use GloVe and word2vec embeddings alongside n-grams for tweet categorization, they still struggle with contextual dependencies. In contrast, transformer-based models like BERT, RoBERTa, and XLNet demonstrate superiority over traditional methods such as GloVe and ELMo \cite{zhou2022victimfinder}. Therefore, in this study, we use the following pre-trained models as baselines against CrisisTransformers \cite{lamsal2024crisistransformers}: MPNet \cite{song2020mpnet}, BERTweet \cite{nguyen2020bertweet}, BERT \cite{devlin-etal-2019-bert}, RoBERTa \cite{liu2019roberta}, XLM-RoBERTa \cite{conneau2019unsupervised}, ALBERT \cite{lan2019albert}, XLNet \cite{yang2019xlnet}, and ELECTRA \cite{clark2020electra}.

Limited work has been done in matching tasks. We identified only two prior studies that attempted to address this problem. Purohit et al. \cite{purohit2014emergency} employed a Gradient Boosted Decision Tree trained on textual similarity and probabilities provided by classifiers as features. They computed textual similarity using cosine similarity on Term Frequency-Inverse Document Frequency (TF-IDF) features of two compared texts. Subsequently, Dutt et al. \cite{dutt2019utilizing} combined textual similarity and normalized spatial distance concerning the country where the disaster occurred. They generated embeddings of resource entities (health, shelter, food, logistics, and cash) using pre-trained word2vec. They performed toponym extraction and used a gazetteer to extract geo-coordinates. The matching score was a linear combination of textual similarity and location proximity score. They implemented their proposed methodology on 1.5k requests and offers in English tweets related to the Nepal and Italy earthquakes. For evaluation, they randomly selected 50 requests and 50 offers tweets from each dataset, and asked human annotators to judge if the matched information was correct.

For the matching tasks, \cite{purohit2014emergency} and \cite{dutt2019utilizing} rely on a frequency-based approach and pre-trained word2vec, respectively, limiting their scope to the English language. However, we demonstrate later in the paper that embedding approaches such as word2vec, GloVe, fastText, etc., are inadequate for the matching tasks. \cite{dutt2019utilizing} utilizes POS taggers and NER tools for information extraction. The identification of resources is based on semantic similarity with a predefined list, potentially restricting the methodology to the completeness and relevance of that list and overlooking context-specific terms. Maintaining such lists for multiple languages is impractical. Additionally, their method linearly combines textual similarity and spatial features, resulting in the matching of offers even from distant regions. Moreover, temporal aspects are disregarded, leading to a neglect of urgent requests that require immediate attention. In this study, we address the aforementioned issues in the matching tasks by utilizing a cross-lingual embedding space to generate state-of-the-art sentence embeddings while considering both spatial and temporal features. Additionally, unlike prior research, we intentionally increase the difficulty of evaluating matching tasks by employing a manually curated set of hard-coded matched requests and offers.

\section{Method}
\label{method}
\begin{figure*}
    \centering
        \includegraphics[width=0.9\textwidth]{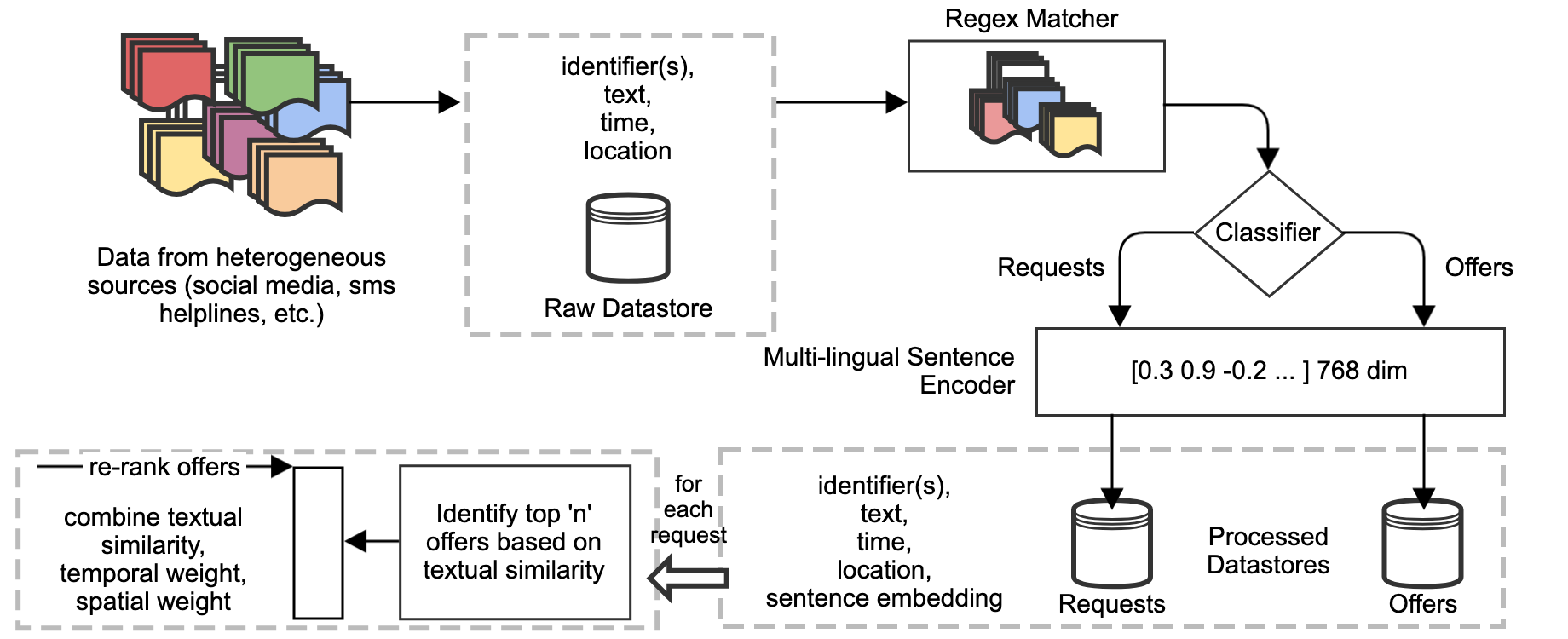}
    \caption{A high-level overview of the proposed approach for matching requests and offers shared on social media.}
    \label{framework}
\end{figure*}

In this section, we outline our approach (CReMa) to matching requests and offers during crises, as depicted in Figure \ref{framework}. The approach involves managing two types of data stores: (i) raw and (ii) processed. The raw data store encompasses all conceivable data collected during crises, including user/account identifiers, texts, time, location, etc. Data is continuously gathered from diverse sources such as social media, web and mobile applications, and crises helplines, and is stored in a database with high writing throughput. 

Following this initial data collection phase, we proceed to identify ``potential'' request and offer candidates by applying a set of regular expressions to both types of texts. Subsequently, we subject each potential request and offer text to two classifiers trained to determine whether the text is exclusively seeking help, offering help, or is irrelevant. This process results in two distinct pools: requests and offers. Additionally, we employ a third classifier to categorize each text into specific categories such as ``money'', ``volunteer'', ``cloth'', ``shelter'', ``medical,'' and ``food''.

Moving forward, we utilize a sentence encoder to encode texts, and this information is stored in the processed data store. Based on the encoded sentence representations, along with the time and location of the help request or offer, we integrate textual, temporal, and spatial features to compute weighted similarity. This enables us to identify the Top-n offers for each request effectively.

Next, we discuss the method in detail.

\subsection{Regular Expressions}
We utilize a set of regular expressions to filter out irrelevant content from the raw data store in order to identify \textit{potential} requests and offers. We improve upon the initial sets proposed in \cite{purohit2014emergency} to develop a more comprehensive set of regular expressions. Initially, a list of regular expressions was manually curated. Subsequently, we matched the texts from datasets provided by \cite{purohit2014emergency} and our own help-offer matching datasets (discussed later in Section \ref{our-datasets}) against the regular expressions. We employed a manual recursive process to add more expressions, incorporating unmatched texts until all the texts from both datasets were matched. The final set had 24 regular expressions. Table \ref{reg-exp} presents the top 5 regular expressions utilized for identifying \textit{potential} requests and offers. These expressions are ranked based on the count of help request and offer texts they successfully match.

Considering the challenge in maintaining a comprehensive set of regular expressions to identify \textit{potential} requests and offers, it becomes crucial to assess its necessity, especially in scenarios involving substantial data influx, such as global events like the COVID-19 pandemic, which generated over 2 million tweets per hour\footnote{\url{https://blog.twitter.com/engineering/en_us/topics/insights/2021/how-we-built-a-data-stream-to-assist-with-covid-19-research}}. These events are unparalleled in scale, and directly subjecting such a massive data stream to classifiers might lead to bottleneck issues. However, our experiments demonstrate that an NVIDIA A100 GPU (80GB) can efficiently process and classify 7.2 million tweets in less than 2 hours, indicating that a single GPU is capable of handling such volumes of data effectively. Consequently, the use of regular expressions as a filtering step becomes optional and can be entirely avoided for smaller-scale events.

\begin{table}
    \centering
\caption{Top 5 regular expressions for identification of \textit{potential} requests and offers.}
    \label{reg-exp}
    \begin{tabular}{l|p{7.6cm}}
    \toprule
       \textbf{SN}  & \textbf{Pattern} \\
       \midrule
1  & {\textbackslash}b(donate \lpipe\ donor \lpipe\ donors \lpipe\ help \lpipe\ fundraising \lpipe\ fundraiser \lpipe\ relief \lpipe\ fund \lpipe\ aiding \lpipe\ volunteer \lpipe\ volunteers \lpipe\ response team \lpipe\ response teams \lpipe\ victim \lpipe\ victims){\textbackslash}b\\
2  & {\textbackslash}b(donate \lpipe\ donating \lpipe\ donation \lpipe\ donations){\textbackslash}b\\
3  & {\textbackslash}b(cloth \lpipe\ clothes \lpipe\ clothing \lpipe\ jersey \lpipe\ sweater \lpipe\ sweaters \lpipe\ vest \lpipe\ vests \lpipe\ jeans \lpipe\ jacket \lpipe\ jackets \lpipe\ blazer \lpipe\ blazers \lpipe\ glove \lpipe\ gloves \lpipe\ blanket \lpipe\ blankets \lpipe\ mask \lpipe\ masks \lpipe\ ppe \lpipe\ sanitizers \lpipe\ supplies){\textbackslash}b\\
4  & {\textbackslash}b(need \lpipe\ needing \lpipe\ inform \lpipe\ informing \lpipe\ looking \lpipe\ sharing \lpipe\ share \lpipe\ offer \lpipe\ offering \lpipe\ providing \lpipe\ seeking \lpipe\ searching){\textbackslash}b.*{\textbackslash}b(supplies \lpipe\ shelter \lpipe\ testing \lpipe\ vaccination \lpipe\ emergency services \lpipe\ help \lpipe\ support \lpipe\ aid \lpipe\ assistance \lpipe\ information \lpipe\ update \lpipe\ updates){\textbackslash}b\\
5  & {\textbackslash}b{\textbackslash}w*{\textbackslash}s*{\textbackslash}b{\textbackslash}?\\
         \bottomrule
    \end{tabular}
\end{table}

\subsection{Classifiers}
\label{classifiers}
Transformer-based \cite{vaswani2017attention} pre-trained models have demonstrated superior performance compared to traditional approaches, significantly advancing the state-of-the-art in natural language processing. The contextualized embeddings generated by these models yield state-of-the-art results across various NLP tasks, including text classification. In this study, we use the following transformer-based pre-trained models as baselines: MPNet, BERTweet, BERT, RoBERTa, XLM-RoBERTa, ALBERT, XLNet, and ELECTRA. In addition to these strong baselines, we also use CrisisTransformers, pre-trained models from our previous study \cite{lamsal2024crisistransformers}, which were trained on a corpus containing more than 15 billion tokens collected from over 30 crisis events that occurred between 2006 and 2023.

\subsubsection{Fine-tuning}
The raw pre-trained models are initially trained for masked language modeling and are designed to be fine-tuned for specific downstream tasks. To fine-tune the pre-trained models for text classification, we added a linear prediction layer to the pooled output. Mean pooling was applied over the token embeddings of an input sequence. Each model was fine-tuned with a maximum of 40 epochs, a batch size of 32, a learning rate of 1e-5, and AdamW as an optimizer with weighted cross-entropy loss. Early stopping was configured with a patience of 8 and a threshold of 0.0001. Fine-tuning was repeated 5 times with different seeds (42, 0, 17, 23, 2024) for each model on 70/10/20 train, validation and test splits. Average performance scores on test data are reported at a 95\% confidence interval.

\subsubsection{Tasks}
This study involves three classification tasks. The first task is identifying whether a text is requesting help. If the text in the first task is not a request for help, the second task determines whether it is offering help. We adopt this cascade methodology (separate binary classifiers) for identifying request and offer texts, aligning with the approach implemented in \cite{purohit2014emergency}. The third task focuses on identifying the type of resource being discussed and is applicable for both requests and offers. Each of these classification tasks require fine-tuning the pre-trained models on three distinct datasets. Detailed information about the datasets is provided in Section \ref{training-datasets}.

\subsubsection{Text preprocessing}
Each tweet in the training datasets was preprocessed: (i) URLs were replaced with ``HTTPURL'' token, (ii) mentions were replaced with an ``@MENTION'' token, (iii) HTML entities were decoded, (iv) newline characters were removed and consecutive whitespaces were replaced with a single space, (v) text encoding issues were corrected for consistency, and (vi) emojis were replaced with their textual representation.

\subsection{Sentence Encoding}
\label{s_text}

We encode request and offer texts to obtain their sentence embeddings. These embedding can then be utilized alongside distance measures like cosine similarity to extract textual similarity. However, out-of-the-box, embeddings generated by pre-trained models like BERT and RoBERTa lack semanticity\footnote{Semantically meaningful sentence embeddings position similar sentences close together in the vector space.} and perform worse than averaging GloVe embeddings \cite{reimers2019sentence}.

In this study, we utilize the multi-lingual sentence encoder from our earlier work \cite{lamsal2023crosslingual} for encoding purposes. The encoder is based on XLM-RoBERTa, which underwent further training as a student in a teacher-student training network with CrisisTransformers' mono-lingual sentence encoder \cite{lamsal2024crisistransformers} as the teacher. The network was trained to make XLM-RoBERTa mimic the embedding space of the mono-lingual sentence encoder using a dataset containing over 128 million multi-lingual parallel sentences, including proceedings from the European Parliament, news stories, bilingual dictionaries, translated subtitles, cross-lingual Wiki data, etc., extending the capabilities of the teacher model to a multi-lingual context.

Given a request text and an offer text, \( T_1 \) and \( T_2 \), the sentence encoder generates embeddings \( E_1 \) and \( E_2 \). Next, we compute the textual similarity:

\begin{equation}
    \text{Textual Similarity (\( S_{\text{text}} \))} = \frac{\mathbf{E}_1 \cdot \mathbf{E}_2}{\|\mathbf{E}_1\| \cdot \|\mathbf{E}_2\|}
    \label{cosine}
\end{equation}

The considered sentence encoder is context-aware regarding location mentions. In a scenario where a plea for O+ blood donation is made, such as ``\textit{We are in need of O+ blood at the Royal Melbourne Hospital. Please help}', the encoder recognizes the ``Melbourne'' context. For instance, when comparing it to an offer like ``\textit{Hey Melbourne, I am ready to donate blood today. Let me know further details}', the sentence encoder yields a similarity score of 0.63. We performed similar evaluations with exact wordings against offers like ``\textit{Hey Victoria, ...}'' (0.55), ``\textit{Hey Aussies, ...}'' (0.59), ``\textit{Hey Sydney, ...}'' (0.53), ``\textit{Hey New York, ...}'' (0.42), ``\textit{Hey Kathmandu, ...}'' (0.41), and ``\textit{Hey New Delhi, ...}'' (0.41). The lower similarity scores for ``\textit{Hey} \{\textit{New York/Kathmandu/New Delhi}\}, ...'' compared to Australian locales indicates the capability of the encoder in understanding geo-locations. Importantly, this scenario holds true for multi-lingual contexts, including when the scripts are non-Latin.

We implement semantic search on embeddings generated by the sentence encoder. The embeddings are based on mean-pooling of tokens with consideration of attention mask.

\subsection{Weighted similarity}
In our approach, we weigh the textual similarity ($S_{text}$) with temporal and spatial weights. Unlike \cite{dutt2019utilizing}, we adopt a non-linear combination of weights. This is crucial because linearly weighing temporal and spatial features could lead to the inclusion of offers from distant regions and time. For example, if a help request for clothes is initiated in ``Melbourne'', offers from Australian locales should be prioritized. Conversely, offers from locations such as New York, Kathmandu, and New Delhi should be disregarded, as assistance from distant places may be better suited for requests in their own nearby regions. Similarly, for a blood request today, assistance offered after a couple of weeks may not be practical or useful.

We emulate human decision-making in our approach by assigning higher weights to offers closer in proximity. For instance, if a help request originates from ``Royal Melbourne Hospital, Parkville, Victoria'', priority is given to offers from nearby areas, gradually extending to the outskirts, other suburbs in Melbourne, regional areas in Victoria, and so on. This prioritization is based on the intuitive notion of what is considered ``near'' and ``far'' in daily life decision-making.

To achieve this, we introduce two crucial hyper-parameters: the maximum allowable waiting time ($\delta_{time}$) and the maximum allowable distance ($\delta_{distance}$). These parameters weigh $S_{\text{text}}$ by incorporating a temporal weight ($W_{\text{time}}$) and a spatial weight ($W_{\text{location}}$). Note that $W_{\text{location}}$ can be computed only when requests and offers are geotagged. Our study assumes that such texts are geotagged; however, acknowledging that this is not always the case, we discuss related limitations later in the discussions.

The overall similarity score ($S_{\text{overall}}$) for a pair of texts is calculated as follows:
\begin{equation}
S_{\text{overall}} = S_{\text{text}} \cdot W_{\text{time}} \cdot W_{\text{location}}
\label{overall-eq}
\end{equation}

Due to the non-linear weighing, offers that do not fall within a temporal and spatial proximity receive an overall similarity score of ``0'' and will not appear as matched offers. We compute $W_{\text{time}}$ and $W_{\text{location}}$ as discussed below:

\subsubsection{Temporal Weight}
Given a request timestamp \( R_{\text{time}} \), an offer timestamp \( Q_{\text{time}} \), and a maximum allowable waiting time \( \delta_{time} \), the time difference \( \Delta t \) is calculated as:
\begin{equation}
    \Delta t = |Q_{\text{time}} - R_{\text{time}}|
\end{equation}
\( W_{\text{time}} \) is then determined using a linear decay function:
\begin{equation}
    W_{\text{time}} = 1 - \min\left(\frac{\Delta t}{\delta_{time}}, 1\right)
    \label{time-decay}
\end{equation}

The timestamps are based on the times when the social media posts were published online.

\subsubsection{Spatial Weight}
Given a request location geo-coordinates \( (R_{\text{lat}}, R_{\text{lon}}) \) and an offer location geo-coordinates \( (Q_{\text{lat}}, Q_{\text{lon}}) \), along with a maximum allowable distance \( \delta_{distance} \), the Haversine formula calculates the great-circle distance (\( d \)) between the two geo-coordinates. The formula involves the following steps:
\begin{equation}
    \begin{aligned}
        a &= \sin^2\left(\frac{\Delta\text{lat}}{2}\right) + \cos(\text{lat}_1) \cdot \cos(\text{lat}_2) \cdot \sin^2\left(\frac{\Delta\text{lon}}{2}\right) \\
        c &= 2 \cdot \text{atan2}\left(\sqrt{a}, \sqrt{1-a}\right) \\
        d &= R \cdot c
    \end{aligned}
\end{equation}
Where \( \Delta\text{lon} = \text{lon}_2 - \text{lon}_1 \), \( \Delta\text{lat} = \text{lat}_2 - \text{lat}_1 \), and \( R \) is the mean radius of the Earth. Now, \( d \) is used to determine \( W_{\text{location}} \):
\begin{equation}
    W_{\text{location}} = 1 - \min\left(\frac{d}{\delta_{distance}}, 1\right)
    \label{location-decay}
\end{equation}

\textit{Significance of the hyper-parameters}: $\delta_{time}$ and $\delta_{location}$ provide flexibility for tailoring the matching algorithm to specific use cases and user preferences, emphasizing the importance of choosing values aligned with application requirements.

$\delta_{time}$ represents the maximum allowable waiting time and controls the temporal proximity between a request and an offer. A smaller $\delta_{time}$ narrows the acceptable time window, favoring real-time or near-real-time matches, while a larger value allows for flexibility in temporal matching. Likewise, $\delta_{distance}$ is maximum allowable distance and controls spatial proximity. A smaller $\delta_{distance}$ confines matches to locations in close physical proximity, ideal for localized services, and a larger $\delta_{distance}$ permits broader geographical matching. Equation \ref{time-decay} ensures that when the time difference $\Delta t$ exceeds $\delta_{time}$, the temporal weight $W_{\text{time}} = 0$, effectively filtering out offers beyond the specified temporal proximity as per Equation \ref{overall-eq}. Similarly, Equation \ref{location-decay} ensures that when $d$ surpasses $\delta_{distance}$, the spatial weight $W_{\text{location}} = 0$, thereby excluding offers beyond the defined spatial proximity.

\subsection{Approximation search}
We implement exhaustive search (brute-force) as a baseline for the vector search task. Additionally, we implement Inverted File Index (IVF), IVF with Product Quantization (PQ) \cite{jegou2010product}, and Hierarchical Navigable Small World (HNSW) \cite{malkov2018efficient} approximation search strategies to study their index/search times and accuracy trade-offs. Since these strategies are widely used and discussed, we do not discuss their theory in this paper.

\subsection{Data}

\subsubsection{Training Dataset}
\label{training-datasets}
To train our classifiers, we utilized the \textit{requests/non-exclusive requests} and \textit{offers/others} datasets introduced in the study \cite{purohit2014emergency}. The released version of the \textit{requests/non-exclusive requests} dataset included 3836 examples, while the \textit{offers/others} dataset comprised 1749 examples. The datasets were released with tweet identifiers and their respective labels. According to Twitter's data redistribution policy, researchers are permitted to share only the identifiers. These identifiers need to be hydrated using Twitter's tweet lookup endpoint to recreate the datasets locally. When tweet identifiers are hydrated, tweets that have been deleted, and tweets from private and suspended accounts cannot be retrieved. Therefore, hydration usually results in a lesser number of tweets than what was originally shared in the form of identifiers. Following hydration, we successfully retrieved 2940 examples from \textit{requests/non-exclusive requests} dataset and 1376 examples from \textit{offers/others} dataset. 450 examples are common to both datasets, labeled as non-exclusive requests in the \textit{requests/non-exclusive requests} dataset and as offers or others in the \textit{offers/others} dataset. Also, \cite{purohit2014emergency} released an additional dataset for the identification of resources across 6 categories: ``money,'' ``volunteer,'' ``cloth', ``shelter', ``medical', and ``food'. The dataset had 3572 labelled examples out of which 2708 were retrieved.

\begin{table}
\centering
\caption{Tweet examples in the \textit{requests/non-exclusive requests} dataset and \textit{offers/others} dataset.}
    \label{examples}
    \begin{tabular}{lp{7cm}}
    \toprule
    \textbf{Label} & \textbf{Example} \\
    \midrule
     request    &  @MENTION: Contribute to the SXSW Sandy Relief Fund to help the Red Cross bring aid to those affected by the hurricane. [HTTPURL] \\
     not-excl    &  Follow @MENTION for updates on how to help with Hurricane Relief!\\
        \midrule
    offer & I'll be bringing 2 boxes \& 2 big bags of clothing \& supplies to help with Hurricane Sandy Relief.\\
    other & Heard my cousin got redeployed with his platoon to help the victims of hurricane sandy, so proud of him. \#nationalguard \#hurricanesandy \\
    \bottomrule
    \end{tabular}
\end{table}

\begin{table*}
\caption{Request and offer examples in \textit{Help-Offer Matching} datasets.}
    \label{examples}
    \centering
    \begin{tabular}{lp{7.9cm}p{7.9cm}}
    \toprule
    \textbf{Dataset}& \textbf{Request} & \textbf{Offer} \\
    \midrule
\multirow{3}{*}{en}&  Need medical supplies in Melbourne's inner city. First aid kits, medications, and basic medical supplies are urgently required for disaster victims.   &  Inner city residents, I'm prepared to provide medical supplies. I have first aid kits, medications, and basic medical supplies. Reach out if you need assistance!                \\
  \midrule
\multirow{3}{*}{es}&  Se informó escasez de sangre de emergencia en el Royal Perth Hospital debido a accidentes recientes. ¿Algún donante disponible?   &  Estoy organizando una campaña de donación de sangre en mi vecindario la próxima semana. Nos aseguraremos de contribuir al Royal Perth Hospital.                \\
\midrule

\multirow{2}{*}{zh-CN} & \begin{CJK*}{UTF8}{gbsn}阿德莱德西郊急需瓶装水。最近的热浪已经耗尽了我们当地的供应。有人可以帮助我们吗？\end{CJK*} & \begin{CJK*}{UTF8}{gbsn}我已准备好向阿德莱德西郊受热浪影响的人们提供瓶装水援助。无需具体要求，只需联系即可！\end{CJK*}\\

\midrule

\multirow{3}{*}{ko} & \begin{CJK*}{UTF8}{mj}시드니의 금융 위기 속에서 새로운 일자리 기회를 확보하기 위해 고군분투하고 있습니다. 취업 시장을 탐색하려면 지침과 리소스가 필요합니다.\end{CJK*} & \begin{CJK*}{UTF8}{mj}우리 조직은 시드니의 경기 침체 기간 동안 취업을 원하는 개인을 지원하기 위해 왔습니다. 우리는 취업 준비 프로그램, 기술 구축 워크숍, 취업 알선 지원을 제공합니다. 맞춤형 지원을 받으려면 문의하세요.\end{CJK*}\\

\midrule

\multirow{9}{*}{random} & [de] Brauche dringend Masken in Fitzroy North. Die örtliche Klinik ist mit einem Mangel konfrontiert. \#Gesundheitskrise & [pt] Fitzroy North, tenho máscaras de sobra. Pronto para entregar em clínicas locais necessitadas. Não é necessária nenhuma solicitação específica. \#HealthcareSupport\\

& [en] Running out of essential supplies due to the lockdown. Need groceries and can't leave home. Any help appreciated. \#LockdownHelp \#Perth & [zh-TW] \begin{CJK*}{UTF8}{gbsn}珀斯的鄰居們，我想讓你們知道我是來幫忙的。如果您在封鎖期間需要雜貨或任何協助，請與我們聯絡。我們在一起！\end{CJK*} \\

& [zh-CN] \begin{CJK*}{UTF8}{gbsn}洪水过后需要临时避难所。悉尼有什么推荐或者优惠吗？ \#悉尼洪水\end{CJK*} & [es] Estoy dispuesto a ofrecer alojamiento temporal a los desplazados por la inundación. Comuníquese y lo coordinaremos. \#Alivio contra las inundaciones de Sídney\\

    \bottomrule
    \end{tabular}
\end{table*}

\subsubsection{Help-Offer Matching Datasets}
\label{our-datasets}

We used GPT-3.5 Turbo for generating synthetic examples in English to simulate help-seeking and offering assistance scenarios for five Australian cities: Sydney, Melbourne, Brisbane, Adelaide, and Perth. The examples were tailored based on probable incidents that have happened in the respective cities. Each example includes the following information: city, help-seeking text language, help-seeking text, help-seeking timestamp, help-seeking location, offer text language, offer text, offer timestamp, and offer location. We manually verified each example for its textual similarity. The initial dataset creation involved the random generation of time and location values. For each example, we considered a maximum 3-day window for timestamps and a maximum 10-kilometre great-circle distance for locations.

After generating the initial dataset, language translation was performed for each help-seeking text and offer text using the Google Cloud Translate API for 15 commonly used languages in Australia. The following languages were considered: Chinese (Simplified), Chinese (Traditional), Arabic, Vietnamese, Greek, Italian, Hindi, Spanish, Korean, Gujarati, Indonesian, Farsi (Persian), French, German, and Portuguese. This step aimed to simulate multi-lingual social media interactions during a crisis. Altogether, 13,264 samples (16 languages * 829) were curated.

To diversify the datasets further, we created 10 random datasets (R1--10) to introduce variability in the distribution of languages by randomly selecting the language for each help-seeking and offer text. 

\begin{itemize}
    \item Help-seeking texts: For each sample, a language was randomly selected from the pool of 16 languages. The corresponding help-seeking text in that language was then assigned to the sample.
    \item Offer texts: Similarly, a separate random selection process was used to assign a language to the offer text for each sample. This language could be the same or different from the one used for the help-seeking text.
\end{itemize}

\begin{figure*}
  \centering
  \subfloat{\includegraphics[width=0.5\linewidth]{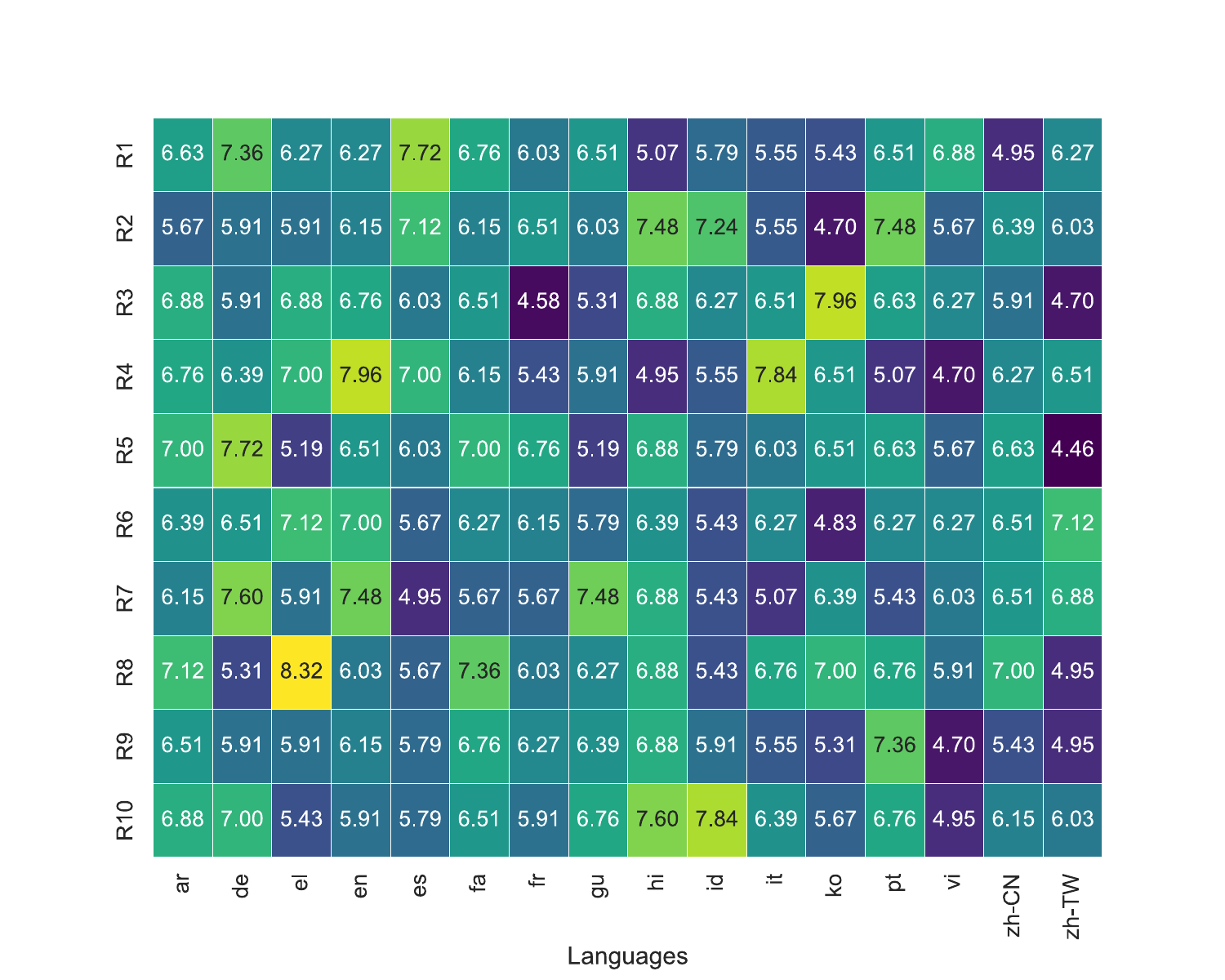}}%
  \hfil
  \subfloat{\includegraphics[width=0.5\linewidth]{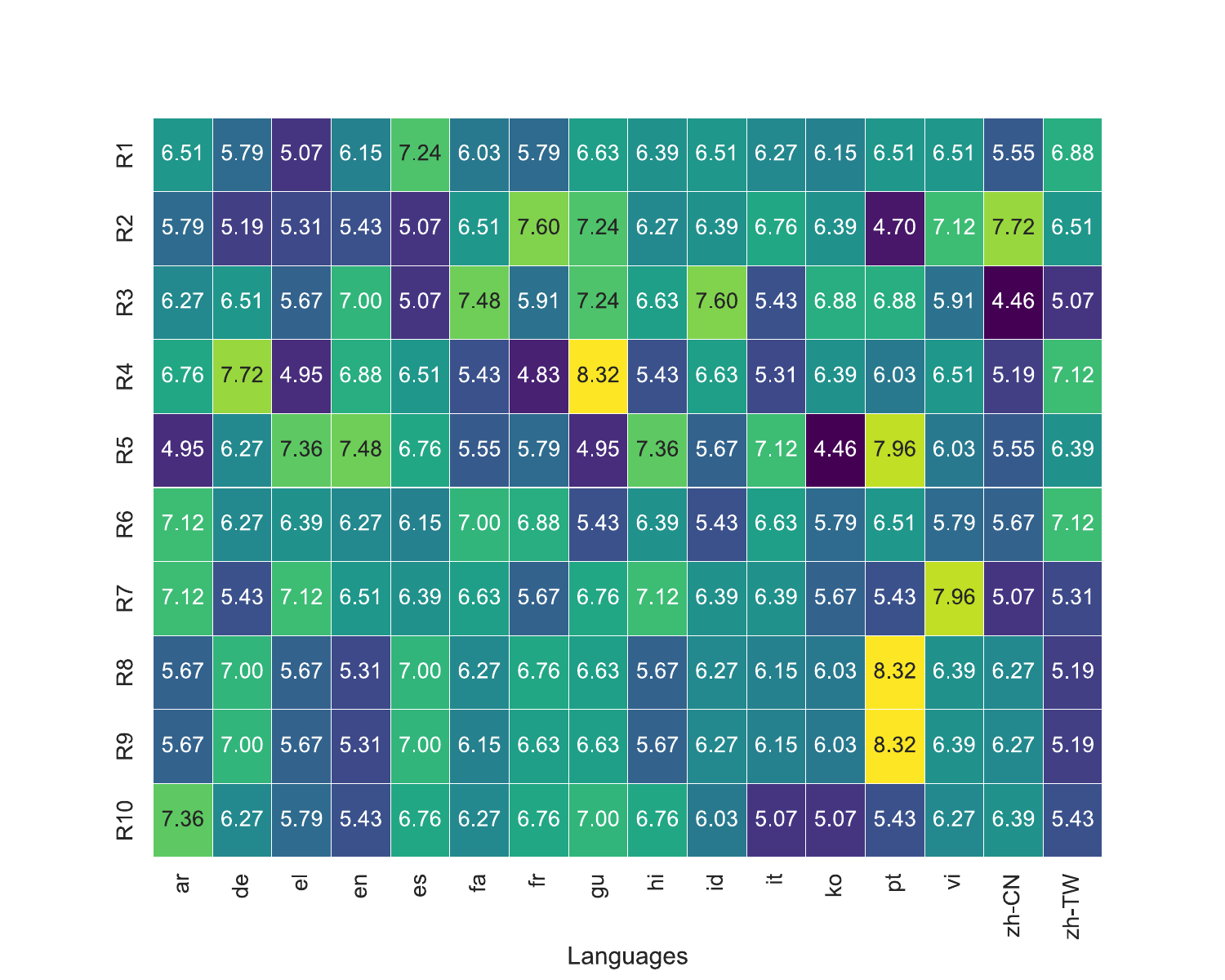}}
  \caption{Illustrations of language distributions (in \%, \textbf{left}: request texts, \textbf{right}: offer texts) across random datasets.}
  \label{lang-dist}
\end{figure*}

The random selection process ensures that each random dataset (R1--R10) features a unique combination of languages with the same number of samples as 16 homogeneous datasets. This approach results in some datasets having higher concentrations of certain languages, while others might have a more even distribution. This step was necessary to study the presence of some languages in the samples and their effect on overall matching performance. Table \ref{examples} provides some requests and offers examples. Figure \ref{lang-dist} shows language distributions across the random datasets.

\subsubsection{Million-scale Conversations}
We analyze a real-world crisis discourse dataset, \textit{MegaGeoCOV Extended} \cite{lamsal2023narrative}, using the classifiers discussed in Section \ref{classifiers}. The dataset is a large-scale multi-lingual collection of COVID-19-related geo-tagged Twitter discourse. Geo-tagging is based on either point location or place information. Since the classifiers are trained on English-only samples, we limit this analysis to English tweets, which total 17.8 million.

\subsection{Evaluation setups}
Following the standard approach \cite{nguyen2020bertweet}, for the classification tasks, we conduct 5 independent runs using the pre-defined seeds and report the average F1 (harmonic mean of precision and recall) with a 95\% confidence interval. While, for the matching tasks, the evaluation is based on Top-$n$ accuracy, where $n$ represents the number of top-ranked results.

\section{Results and discussions}
\label{result}
\subsection{Classification}

\begin{table}
\caption{Classification performance of different transformer-based models. CrisisTransformers offers 8 pre-trained models; here, we list the top 3 classifiers.}
\label{clf-results}
  \centering
  \subfloat[Request -- non-exclusive request classification.]{
\begin{tabular}{ll}
\toprule
                                           & \textbf{F1$_{Avg.}$ (95\% conf. int.)} \\ \midrule
\textbf{microsoft/mpnet-base}              & 0.8940 ±0.0070 (±0.79\%)               \\
\textbf{vinai/bertweet-covid19-base-cased} & 0.8918 ±0.0059 (±0.66\%)               \\
\textbf{bert-base-cased}                   & 0.8799 ±0.0055 (±0.63\%)               \\
\textbf{roberta-base}                      & 0.8892 ±0.0090 (±1.01\%)               \\
\textbf{xlm-roberta-base}                  & 0.8833 ±0.0085 (±0.97\%)               \\
\textbf{albert-base-v2}                    & 0.8594 ±0.0067 (±0.78\%)               \\
\textbf{xlnet-base-cased}                  & 0.8781 ±0.0064 (±0.74\%)               \\
\textbf{google/electra-base-discriminator} & 0.8804 ±0.0030 (±0.34\%)               \\ \midrule
\textbf{crisistransformers/CT-M1-Complete} & \textbf{0.8967 ±0.0076 (±0.85\%)}      \\
\textbf{crisistransformers/CT-M2-Complete} & 0.8952 ±0.0007 (±0.08\%)               \\
\textbf{crisistransformers/CT-M1-BestLoss} & 0.8945 ±0.0095 (±1.07\%)               \\ \bottomrule
\end{tabular}

  }

  \vspace{0.5em}

  \subfloat[Offer -- other classification.]{
\begin{tabular}{ll}
\toprule
                            & \textbf{F1$_{Avg.}$ (95\% conf. int.)} \\ \midrule
\textbf{microsoft/mpnet-base}              & 0.7277 ±0.0319 (±4.38\%)               \\
\textbf{vinai/bertweet-covid19-base-cased} & 0.7061 ±0.0320 (±4.53\%)               \\
\textbf{bert-base-cased}                   & 0.6985 ±0.0309 (±4.42\%)               \\
\textbf{roberta-base}                      & 0.7348 ±0.0173 (±2.36\%)               \\
\textbf{xlm-roberta-base}                  & 0.6670 ±0.0861 (±12.92\%)              \\
\textbf{albert-base-v2}                    & 0.6655 ±0.0413 (±6.20\%)               \\
\textbf{xlnet-base-cased}                  & 0.7192 ±0.0260 (±3.62\%)               \\
\textbf{google/electra-base-discriminator} & 0.7136 ±0.0443 (±6.20\%)               \\ \midrule
\textbf{crisistransformers/CT-M1-Complete} & \textbf{0.7615 ±0.0450 (±5.91\%)}      \\
\textbf{crisistransformers/CT-M2-BestLoss} & 0.7497 ±0.0402 (±5.36\%)               \\
\textbf{crisistransformers/CT-M1-BestLoss} & 0.7401 ±0.0260 (±3.51\%)               \\ \bottomrule
\end{tabular}
  }

  \vspace{0.5em}

  \subfloat[Resources classification.]{
\begin{tabular}{ll}
\toprule
                           & \textbf{F1$_{Avg.}$ (95\% conf. int.)} \\ \midrule
\textbf{microsoft/mpnet-base}              & 0.9437 ±0.0127 (±1.34\%)               \\
\textbf{vinai/bertweet-covid19-base-cased} & 0.938 ±0.0134 (±1.43\%)                \\
\textbf{bert-base-cased}                   & 0.9306 ±0.0162 (±1.75\%)               \\
\textbf{roberta-base}                      & 0.9413 ±0.0153 (±1.63\%)               \\
\textbf{xlm-roberta-base}                  & 0.9295 ±0.0183 (±1.97\%)               \\
\textbf{albert-base-v2}                    & 0.9339 ±0.0157 (±1.68\%)               \\
\textbf{xlnet-base-cased}                  & 0.9289 ±0.0232 (±2.50\%)               \\
\textbf{google/electra-base-discriminator} & 0.9337 ±0.0119 (±1.28\%)               \\ \midrule
\textbf{crisistransformers/CT-M1-Complete} & \textbf{0.9475 ±0.0168 (±1.77\%)}      \\
\textbf{crisistransformers/CT-M1-BestLoss} & 0.9413 ±0.0186 (±1.98\%)               \\
\textbf{crisistransformers/CT-M3-Complete} & 0.9417 ±0.0177 (±1.87\%)               \\ \bottomrule
\end{tabular}
  }
\end{table}

As discussed earlier, we trained MPNet, BERTweet, BERT, RoBERTa, XLM-RoBERTa, ALBERT, XLNet, and ELECTRA, and 8 pre-trained CrisisTransformers on \textit{requests/non-exclusive requests}, \textit{offers/others}, and \textit{resources} datasets. Table \ref{clf-results} presents the classification performance of different models on three different tasks: \textit{Exclusively request/other} classification, \textit{Offer/other} classification, and \textit{Resources} classification. The evaluation metrics include F1 across five seeds and average F1 with 95\% confidence intervals.

While RoBERTa and MPNet show competitive performance amongst the baselines, the CrisisTransformers family, particularly the \textit{CT-M1-Complete} variant, surpasses them across all three classification tasks, confirming the efficacy of CrisisTransformers in crisis-related text classification. In the first, second, and third classification tasks, \textit{CT-M1-Complete} consistently achieved the highest average F1 scores of 0.8967 (±0.85\%), 0.7615 (±5.91\%), and 0.9475 (±1.77\%), respectively. The superior performance of CrisisTransformers can be attributed to their pre-training on a billion-scale crisis-specific corpus, consistently outperforming strong baselines like BERT and RoBERTa, which were pre-trained on general domain texts. These findings, combined with insights from 18 crisis-specific datasets \cite{lamsal2024crisistransformers}, show the robustness and effectiveness of CrisisTransformers' models, making them reliable choices for designing classifiers in the field of crisis informatics. Table \ref{final-models} provides the F1, Recall and Precision of the final set of classifiers considered in this study.

\begin{table}[h!]
    \centering
    \caption{Final set of classifiers.}
    \label{final-models}
    \begin{tabular}{llll}
    \toprule
    \textbf{Classification task} & \textbf{F1} & \textbf{Precision} & \textbf{Recall} \\
    \midrule
      Requests -- non-exclusive requests   & 0.9169 & 0.9203 & 0.9146\\
      Offers -- others   & 0.8001 & 0.7920 & 0.8089\\
      Resources & 0.9767 & 0.9796 & 0.9743 \\
      \bottomrule
    \end{tabular}
\end{table}

\subsection{Requests/Offers Matching}

Given the abundant resources available for training embedding models in English, we conducted an initial benchmarking of our multi-lingual sentence encoder against several established models, including word2vec, GloVe, fastText, InferSent, Universal Sentence Encoder, and Sentence Transformers (SBERT). The matching tasks were done solely based on textual features, i.e., cosine similarity of embeddings. The results, presented in Table \ref{en-matching-accuracy}, show that our CrisisTransformers' mono-lingual model surpasses all baselines, with the multi-lingual version closely following. Since the multi-lingual model was trained to replicate the embedding space of the mono-lingual model using over 128 million translation pairs, here-on-after, in the study, we employ the multi-lingual model for matching tasks in multi-lingual scenarios, thus establishing a benchmark for future studies.

We compare our approach, referred to as ``TTS'' (\textbf{T}extual \textbf{T}emporal \textbf{S}patial), with the baselines ``T'' and ``TS'', where ``T'' relies on textual features and ``TS'' relies on linear weighing of textual and spatial features. We experimented with the following proximities wherever applicable: $\delta_{time} = \{30, 90, 180\}$ (days), $\delta_{location} = \{10,20,30\}$ (kms), and nearest neighbor ($K$) = \{25, 50, 100\}. Due to the limited size of the synthetic dataset, we conducted an exhaustive search to identify the Top-$n$ nearest offers for each request. This exhaustive search serves as a benchmark for evaluating approximation search strategies. It enables us to investigate the feasibility and accuracy trade-offs of alternative approaches that do not require exploring the entire search space in real-time. Additionally, the resource classifier can categorize tweets into various classes like ``money'', ``volunteer'', etc. This classification can help narrow down the search space during matching tasks. However, in this study, we do not use this information to limit the search space due to the small size of the dataset.

\begin{table}
    \centering
        \caption{Performance of various embedding models on matching tasks on English texts.}
    \label{en-matching-accuracy}
    \begin{tabular}{c|c|c|c}
    \toprule
    \multirow{2}{*}{\textbf{Embedding model}} & \multicolumn{3}{c}{\textbf{Matching accuracy}} \\
     & k=25 & k=50 & k=100 \\
    \midrule
      word2vec (avg.)  & 0.18 & 0.25 & 0.31\\
       GloVe (avg.) & 0.12 & 0.16 & 0.20\\
       fastText (avg.) & 0.05 & 0.08 & 0.09 \\
       InferSent (GloVe) & 0.21 & 0.30 & 0.32\\
       Universal Sentence Encoder & 0.38 & 0.49 & 0.54\\
       SBERT (all-mpnet-base-v2) &  \underline{0.43} & 0.57 & 0.62\\
       \midrule
       CrisisTransformers (mono-lingual) & \textbf{0.45} & \textbf{0.60} & \textbf{0.67}\\
       CrisisTransformers (multi-lingual) & \underline{0.43} & \underline{0.58} & \underline{0.64} \\
       \bottomrule
    \end{tabular}
\end{table}

\begin{table*}
    \centering
\caption{Results from the matching tasks for $\delta_{time} =30$ and $\delta_{location} =10$. ``T'' uses textual features, ``TTS'' uses textual, temporal and spatial features combined non-linearly.}
\label{matching-results}

\begin{tabular}{lcccccccccccc}
\toprule
Accuracy $\rightarrow$         & \textbf{Top-1 (T)}                                    & \multicolumn{3}{c}{\textbf{Top-1 (TTS)}}                                                                                                                            & \textbf{Top-2 (T)}                                    & \multicolumn{3}{c}{\textbf{Top-2 (TTS)}}                                                                                                                            & \textbf{Top-3 (T)}                                    & \multicolumn{3}{c}{\textbf{Top-3 (TTS)}}                                                                                                                            \\
\midrule
Lang $\downarrow$ K $\rightarrow$       & 25/50/100                                                   & 25                                                   & 50                                                   & 100                                                  & 25/50/100                                                   & 25                                                   & 50                                                   & 100                                                  & 25/50/100                                                   & 25                                                   & 50                                                   & 100                                                  \\
\midrule
\textbf{en}       & {\cellcolor[HTML]{365C8D}} \color[HTML]{F1F1F1} 0.43 & {\cellcolor[HTML]{9BD93C}} \color[HTML]{000000} 0.8  & {\cellcolor[HTML]{A5DB36}} \color[HTML]{000000} 0.81 & {\cellcolor[HTML]{ADDC30}} \color[HTML]{000000} 0.82 & {\cellcolor[HTML]{20938C}} \color[HTML]{F1F1F1} 0.58 & {\cellcolor[HTML]{C2DF23}} \color[HTML]{000000} 0.84 & {\cellcolor[HTML]{D8E219}} \color[HTML]{000000} 0.86 & {\cellcolor[HTML]{ECE51B}} \color[HTML]{000000} 0.88 & {\cellcolor[HTML]{24AA83}} \color[HTML]{F1F1F1} 0.64 & {\cellcolor[HTML]{CDE11D}} \color[HTML]{000000} 0.85 & {\cellcolor[HTML]{ECE51B}} \color[HTML]{000000} 0.88 & {\cellcolor[HTML]{FDE725}} \color[HTML]{000000} 0.9  \\
\midrule
\textbf{zh-CN}    & {\cellcolor[HTML]{443983}} \color[HTML]{F1F1F1} 0.35 & {\cellcolor[HTML]{5AC864}} \color[HTML]{000000} 0.73 & {\cellcolor[HTML]{69CD5B}} \color[HTML]{000000} 0.75 & {\cellcolor[HTML]{7CD250}} \color[HTML]{000000} 0.77 & {\cellcolor[HTML]{2C728E}} \color[HTML]{F1F1F1} 0.49 & {\cellcolor[HTML]{73D056}} \color[HTML]{000000} 0.76 & {\cellcolor[HTML]{86D549}} \color[HTML]{000000} 0.78 & {\cellcolor[HTML]{A5DB36}} \color[HTML]{000000} 0.81 & {\cellcolor[HTML]{228D8D}} \color[HTML]{F1F1F1} 0.56 & {\cellcolor[HTML]{7CD250}} \color[HTML]{000000} 0.77 & {\cellcolor[HTML]{9BD93C}} \color[HTML]{000000} 0.8  & {\cellcolor[HTML]{B8DE29}} \color[HTML]{000000} 0.83 \\
\textbf{zh-TW}    & {\cellcolor[HTML]{443983}} \color[HTML]{F1F1F1} 0.35 & {\cellcolor[HTML]{4AC16D}} \color[HTML]{000000} 0.71 & {\cellcolor[HTML]{60CA60}} \color[HTML]{000000} 0.74 & {\cellcolor[HTML]{73D056}} \color[HTML]{000000} 0.76 & {\cellcolor[HTML]{31688E}} \color[HTML]{F1F1F1} 0.46 & {\cellcolor[HTML]{60CA60}} \color[HTML]{000000} 0.74 & {\cellcolor[HTML]{86D549}} \color[HTML]{000000} 0.78 & {\cellcolor[HTML]{9BD93C}} \color[HTML]{000000} 0.8  & {\cellcolor[HTML]{25858E}} \color[HTML]{F1F1F1} 0.54 & {\cellcolor[HTML]{69CD5B}} \color[HTML]{000000} 0.75 & {\cellcolor[HTML]{90D743}} \color[HTML]{000000} 0.79 & {\cellcolor[HTML]{ADDC30}} \color[HTML]{000000} 0.82 \\
\textbf{ar}       & {\cellcolor[HTML]{482677}} \color[HTML]{F1F1F1} 0.31 & {\cellcolor[HTML]{2FB47C}} \color[HTML]{F1F1F1} 0.67 & {\cellcolor[HTML]{3BBB75}} \color[HTML]{F1F1F1} 0.69 & {\cellcolor[HTML]{4AC16D}} \color[HTML]{000000} 0.71 & {\cellcolor[HTML]{365C8D}} \color[HTML]{F1F1F1} 0.43 & {\cellcolor[HTML]{4AC16D}} \color[HTML]{000000} 0.71 & {\cellcolor[HTML]{60CA60}} \color[HTML]{000000} 0.74 & {\cellcolor[HTML]{73D056}} \color[HTML]{000000} 0.76 & {\cellcolor[HTML]{2C728E}} \color[HTML]{F1F1F1} 0.49 & {\cellcolor[HTML]{52C569}} \color[HTML]{000000} 0.72 & {\cellcolor[HTML]{69CD5B}} \color[HTML]{000000} 0.75 & {\cellcolor[HTML]{86D549}} \color[HTML]{000000} 0.78 \\
\textbf{vi}       & {\cellcolor[HTML]{3F4788}} \color[HTML]{F1F1F1} 0.38 & {\cellcolor[HTML]{60CA60}} \color[HTML]{000000} 0.74 & {\cellcolor[HTML]{69CD5B}} \color[HTML]{000000} 0.75 & {\cellcolor[HTML]{7CD250}} \color[HTML]{000000} 0.77 & {\cellcolor[HTML]{2C728E}} \color[HTML]{F1F1F1} 0.49 & {\cellcolor[HTML]{86D549}} \color[HTML]{000000} 0.78 & {\cellcolor[HTML]{9BD93C}} \color[HTML]{000000} 0.8  & {\cellcolor[HTML]{B8DE29}} \color[HTML]{000000} 0.83 & {\cellcolor[HTML]{23898E}} \color[HTML]{F1F1F1} 0.55 & {\cellcolor[HTML]{90D743}} \color[HTML]{000000} 0.79 & {\cellcolor[HTML]{ADDC30}} \color[HTML]{000000} 0.82 & {\cellcolor[HTML]{CDE11D}} \color[HTML]{000000} 0.85 \\
\textbf{el}       & {\cellcolor[HTML]{433E85}} \color[HTML]{F1F1F1} 0.36 & {\cellcolor[HTML]{60CA60}} \color[HTML]{000000} 0.74 & {\cellcolor[HTML]{7CD250}} \color[HTML]{000000} 0.77 & {\cellcolor[HTML]{90D743}} \color[HTML]{000000} 0.79 & {\cellcolor[HTML]{2A768E}} \color[HTML]{F1F1F1} 0.5  & {\cellcolor[HTML]{86D549}} \color[HTML]{000000} 0.78 & {\cellcolor[HTML]{A5DB36}} \color[HTML]{000000} 0.81 & {\cellcolor[HTML]{C2DF23}} \color[HTML]{000000} 0.84 & {\cellcolor[HTML]{228D8D}} \color[HTML]{F1F1F1} 0.56 & {\cellcolor[HTML]{90D743}} \color[HTML]{000000} 0.79 & {\cellcolor[HTML]{B8DE29}} \color[HTML]{000000} 0.83 & {\cellcolor[HTML]{E2E418}} \color[HTML]{000000} 0.87 \\
\textbf{it}       & {\cellcolor[HTML]{433E85}} \color[HTML]{F1F1F1} 0.36 & {\cellcolor[HTML]{73D056}} \color[HTML]{000000} 0.76 & {\cellcolor[HTML]{86D549}} \color[HTML]{000000} 0.78 & {\cellcolor[HTML]{9BD93C}} \color[HTML]{000000} 0.8  & {\cellcolor[HTML]{2D708E}} \color[HTML]{F1F1F1} 0.48 & {\cellcolor[HTML]{9BD93C}} \color[HTML]{000000} 0.8  & {\cellcolor[HTML]{B8DE29}} \color[HTML]{000000} 0.83 & {\cellcolor[HTML]{CDE11D}} \color[HTML]{000000} 0.85 & {\cellcolor[HTML]{228D8D}} \color[HTML]{F1F1F1} 0.56 & {\cellcolor[HTML]{A5DB36}} \color[HTML]{000000} 0.81 & {\cellcolor[HTML]{C2DF23}} \color[HTML]{000000} 0.84 & {\cellcolor[HTML]{D8E219}} \color[HTML]{000000} 0.87 \\
\textbf{hi}       & {\cellcolor[HTML]{460B5E}} \color[HTML]{F1F1F1} 0.26 & {\cellcolor[HTML]{1F978B}} \color[HTML]{F1F1F1} 0.59 & {\cellcolor[HTML]{1F9F88}} \color[HTML]{F1F1F1} 0.61 & {\cellcolor[HTML]{1FA287}} \color[HTML]{F1F1F1} 0.62 & {\cellcolor[HTML]{414287}} \color[HTML]{F1F1F1} 0.37 & {\cellcolor[HTML]{1FA287}} \color[HTML]{F1F1F1} 0.62 & {\cellcolor[HTML]{27AD81}} \color[HTML]{F1F1F1} 0.65 & {\cellcolor[HTML]{2FB47C}} \color[HTML]{F1F1F1} 0.67 & {\cellcolor[HTML]{32648E}} \color[HTML]{F1F1F1} 0.45 & {\cellcolor[HTML]{21A685}} \color[HTML]{F1F1F1} 0.63 & {\cellcolor[HTML]{2FB47C}} \color[HTML]{F1F1F1} 0.67 & {\cellcolor[HTML]{3BBB75}} \color[HTML]{F1F1F1} 0.69 \\
\textbf{es}       & {\cellcolor[HTML]{3F4788}} \color[HTML]{F1F1F1} 0.38 & {\cellcolor[HTML]{73D056}} \color[HTML]{000000} 0.76 & {\cellcolor[HTML]{90D743}} \color[HTML]{000000} 0.79 & {\cellcolor[HTML]{9BD93C}} \color[HTML]{000000} 0.8  & {\cellcolor[HTML]{277E8E}} \color[HTML]{F1F1F1} 0.52 & {\cellcolor[HTML]{9BD93C}} \color[HTML]{000000} 0.8  & {\cellcolor[HTML]{B8DE29}} \color[HTML]{000000} 0.83 & {\cellcolor[HTML]{CDE11D}} \color[HTML]{000000} 0.85 & {\cellcolor[HTML]{20938C}} \color[HTML]{F1F1F1} 0.58 & {\cellcolor[HTML]{9BD93C}} \color[HTML]{000000} 0.8  & {\cellcolor[HTML]{C2DF23}} \color[HTML]{000000} 0.84 & {\cellcolor[HTML]{D8E219}} \color[HTML]{000000} 0.86 \\
\textbf{ko}       & {\cellcolor[HTML]{472F7D}} \color[HTML]{F1F1F1} 0.33 & {\cellcolor[HTML]{42BE71}} \color[HTML]{F1F1F1} 0.7  & {\cellcolor[HTML]{5AC864}} \color[HTML]{000000} 0.73 & {\cellcolor[HTML]{60CA60}} \color[HTML]{000000} 0.74 & {\cellcolor[HTML]{365C8D}} \color[HTML]{F1F1F1} 0.43 & {\cellcolor[HTML]{69CD5B}} \color[HTML]{000000} 0.75 & {\cellcolor[HTML]{90D743}} \color[HTML]{000000} 0.79 & {\cellcolor[HTML]{A5DB36}} \color[HTML]{000000} 0.81 & {\cellcolor[HTML]{2A768E}} \color[HTML]{F1F1F1} 0.5  & {\cellcolor[HTML]{73D056}} \color[HTML]{000000} 0.76 & {\cellcolor[HTML]{9BD93C}} \color[HTML]{000000} 0.8  & {\cellcolor[HTML]{B8DE29}} \color[HTML]{000000} 0.83 \\
\textbf{gu}       & {\cellcolor[HTML]{471164}} \color[HTML]{F1F1F1} 0.27 & {\cellcolor[HTML]{25858E}} \color[HTML]{F1F1F1} 0.54 & {\cellcolor[HTML]{23898E}} \color[HTML]{F1F1F1} 0.55 & {\cellcolor[HTML]{228D8D}} \color[HTML]{F1F1F1} 0.56 & {\cellcolor[HTML]{433E85}} \color[HTML]{F1F1F1} 0.36 & {\cellcolor[HTML]{21908D}} \color[HTML]{F1F1F1} 0.57 & {\cellcolor[HTML]{1F978B}} \color[HTML]{F1F1F1} 0.59 & {\cellcolor[HTML]{1E9B8A}} \color[HTML]{F1F1F1} 0.6  & {\cellcolor[HTML]{365C8D}} \color[HTML]{F1F1F1} 0.43 & {\cellcolor[HTML]{20938C}} \color[HTML]{F1F1F1} 0.58 & {\cellcolor[HTML]{1E9B8A}} \color[HTML]{F1F1F1} 0.6  & {\cellcolor[HTML]{1F9F88}} \color[HTML]{F1F1F1} 0.61 \\
\textbf{id}       & {\cellcolor[HTML]{3E4C8A}} \color[HTML]{F1F1F1} 0.39 & {\cellcolor[HTML]{73D056}} \color[HTML]{000000} 0.76 & {\cellcolor[HTML]{86D549}} \color[HTML]{000000} 0.78 & {\cellcolor[HTML]{90D743}} \color[HTML]{000000} 0.79 & {\cellcolor[HTML]{277E8E}} \color[HTML]{F1F1F1} 0.52 & {\cellcolor[HTML]{A5DB36}} \color[HTML]{000000} 0.81 & {\cellcolor[HTML]{C2DF23}} \color[HTML]{000000} 0.84 & {\cellcolor[HTML]{E2E418}} \color[HTML]{000000} 0.87 & {\cellcolor[HTML]{20938C}} \color[HTML]{F1F1F1} 0.58 & {\cellcolor[HTML]{ADDC30}} \color[HTML]{000000} 0.82 & {\cellcolor[HTML]{CDE11D}} \color[HTML]{000000} 0.85 & {\cellcolor[HTML]{ECE51B}} \color[HTML]{000000} 0.88 \\
\textbf{fa}       & {\cellcolor[HTML]{482173}} \color[HTML]{F1F1F1} 0.3  & {\cellcolor[HTML]{27AD81}} \color[HTML]{F1F1F1} 0.65 & {\cellcolor[HTML]{2FB47C}} \color[HTML]{F1F1F1} 0.67 & {\cellcolor[HTML]{35B779}} \color[HTML]{F1F1F1} 0.68 & {\cellcolor[HTML]{3C508B}} \color[HTML]{F1F1F1} 0.4  & {\cellcolor[HTML]{2FB47C}} \color[HTML]{F1F1F1} 0.67 & {\cellcolor[HTML]{42BE71}} \color[HTML]{F1F1F1} 0.7  & {\cellcolor[HTML]{4AC16D}} \color[HTML]{000000} 0.71 & {\cellcolor[HTML]{2D708E}} \color[HTML]{F1F1F1} 0.48 & {\cellcolor[HTML]{35B779}} \color[HTML]{F1F1F1} 0.68 & {\cellcolor[HTML]{4AC16D}} \color[HTML]{000000} 0.71 & {\cellcolor[HTML]{52C569}} \color[HTML]{000000} 0.72 \\
\textbf{fr}       & {\cellcolor[HTML]{463480}} \color[HTML]{F1F1F1} 0.34 & {\cellcolor[HTML]{60CA60}} \color[HTML]{000000} 0.74 & {\cellcolor[HTML]{7CD250}} \color[HTML]{000000} 0.77 & {\cellcolor[HTML]{90D743}} \color[HTML]{000000} 0.79 & {\cellcolor[HTML]{2D708E}} \color[HTML]{F1F1F1} 0.48 & {\cellcolor[HTML]{86D549}} \color[HTML]{000000} 0.78 & {\cellcolor[HTML]{ADDC30}} \color[HTML]{000000} 0.82 & {\cellcolor[HTML]{C2DF23}} \color[HTML]{000000} 0.84 & {\cellcolor[HTML]{25858E}} \color[HTML]{F1F1F1} 0.54 & {\cellcolor[HTML]{90D743}} \color[HTML]{000000} 0.79 & {\cellcolor[HTML]{B8DE29}} \color[HTML]{000000} 0.83 & {\cellcolor[HTML]{D8E219}} \color[HTML]{000000} 0.86 \\
\textbf{de}       & {\cellcolor[HTML]{3F4788}} \color[HTML]{F1F1F1} 0.38 & {\cellcolor[HTML]{73D056}} \color[HTML]{000000} 0.76 & {\cellcolor[HTML]{86D549}} \color[HTML]{000000} 0.78 & {\cellcolor[HTML]{9BD93C}} \color[HTML]{000000} 0.8  & {\cellcolor[HTML]{277E8E}} \color[HTML]{F1F1F1} 0.52 & {\cellcolor[HTML]{9BD93C}} \color[HTML]{000000} 0.8  & {\cellcolor[HTML]{B8DE29}} \color[HTML]{000000} 0.83 & {\cellcolor[HTML]{C2DF23}} \color[HTML]{000000} 0.84 & {\cellcolor[HTML]{20938C}} \color[HTML]{F1F1F1} 0.58 & {\cellcolor[HTML]{A5DB36}} \color[HTML]{000000} 0.81 & {\cellcolor[HTML]{C2DF23}} \color[HTML]{000000} 0.84 & {\cellcolor[HTML]{E2E418}} \color[HTML]{000000} 0.87 \\
\textbf{pt}       & {\cellcolor[HTML]{3E4C8A}} \color[HTML]{F1F1F1} 0.39 & {\cellcolor[HTML]{7CD250}} \color[HTML]{000000} 0.77 & {\cellcolor[HTML]{86D549}} \color[HTML]{000000} 0.78 & {\cellcolor[HTML]{9BD93C}} \color[HTML]{000000} 0.8  & {\cellcolor[HTML]{26828E}} \color[HTML]{F1F1F1} 0.53 & {\cellcolor[HTML]{9BD93C}} \color[HTML]{000000} 0.8  & {\cellcolor[HTML]{C2DF23}} \color[HTML]{000000} 0.84 & {\cellcolor[HTML]{D8E219}} \color[HTML]{000000} 0.86 & {\cellcolor[HTML]{1F978B}} \color[HTML]{F1F1F1} 0.59 & {\cellcolor[HTML]{ADDC30}} \color[HTML]{000000} 0.82 & {\cellcolor[HTML]{CDE11D}} \color[HTML]{000000} 0.85 & {\cellcolor[HTML]{ECE51B}} \color[HTML]{000000} 0.88 \\
\midrule
\textbf{R1}       & {\cellcolor[HTML]{460B5E}} \color[HTML]{F1F1F1} 0.26 & {\cellcolor[HTML]{1FA287}} \color[HTML]{F1F1F1} 0.62 & {\cellcolor[HTML]{24AA83}} \color[HTML]{F1F1F1} 0.64 & {\cellcolor[HTML]{27AD81}} \color[HTML]{F1F1F1} 0.65 & {\cellcolor[HTML]{3F4788}} \color[HTML]{F1F1F1} 0.38 & {\cellcolor[HTML]{2AB07F}} \color[HTML]{F1F1F1} 0.66 & {\cellcolor[HTML]{3BBB75}} \color[HTML]{F1F1F1} 0.69 & {\cellcolor[HTML]{42BE71}} \color[HTML]{F1F1F1} 0.7  & {\cellcolor[HTML]{34608D}} \color[HTML]{F1F1F1} 0.44 & {\cellcolor[HTML]{2AB07F}} \color[HTML]{F1F1F1} 0.66 & {\cellcolor[HTML]{3BBB75}} \color[HTML]{F1F1F1} 0.69 & {\cellcolor[HTML]{4AC16D}} \color[HTML]{000000} 0.71 \\
\textbf{R2}       & {\cellcolor[HTML]{481769}} \color[HTML]{F1F1F1} 0.28 & {\cellcolor[HTML]{20938C}} \color[HTML]{F1F1F1} 0.58 & {\cellcolor[HTML]{1F9F88}} \color[HTML]{F1F1F1} 0.61 & {\cellcolor[HTML]{21A685}} \color[HTML]{F1F1F1} 0.63 & {\cellcolor[HTML]{414287}} \color[HTML]{F1F1F1} 0.37 & {\cellcolor[HTML]{21A685}} \color[HTML]{F1F1F1} 0.63 & {\cellcolor[HTML]{2AB07F}} \color[HTML]{F1F1F1} 0.66 & {\cellcolor[HTML]{3BBB75}} \color[HTML]{F1F1F1} 0.69 & {\cellcolor[HTML]{365C8D}} \color[HTML]{F1F1F1} 0.43 & {\cellcolor[HTML]{27AD81}} \color[HTML]{F1F1F1} 0.65 & {\cellcolor[HTML]{35B779}} \color[HTML]{F1F1F1} 0.68 & {\cellcolor[HTML]{4AC16D}} \color[HTML]{000000} 0.71 \\
\textbf{R3}       & {\cellcolor[HTML]{471164}} \color[HTML]{F1F1F1} 0.27 & {\cellcolor[HTML]{20938C}} \color[HTML]{F1F1F1} 0.58 & {\cellcolor[HTML]{1FA287}} \color[HTML]{F1F1F1} 0.62 & {\cellcolor[HTML]{21A685}} \color[HTML]{F1F1F1} 0.63 & {\cellcolor[HTML]{414287}} \color[HTML]{F1F1F1} 0.37 & {\cellcolor[HTML]{1FA287}} \color[HTML]{F1F1F1} 0.62 & {\cellcolor[HTML]{2FB47C}} \color[HTML]{F1F1F1} 0.67 & {\cellcolor[HTML]{3BBB75}} \color[HTML]{F1F1F1} 0.69 & {\cellcolor[HTML]{34608D}} \color[HTML]{F1F1F1} 0.44 & {\cellcolor[HTML]{21A685}} \color[HTML]{F1F1F1} 0.63 & {\cellcolor[HTML]{3BBB75}} \color[HTML]{F1F1F1} 0.69 & {\cellcolor[HTML]{4AC16D}} \color[HTML]{000000} 0.71 \\
\textbf{R4}       & {\cellcolor[HTML]{450559}} \color[HTML]{F1F1F1} 0.25 & {\cellcolor[HTML]{1E9B8A}} \color[HTML]{F1F1F1} 0.6  & {\cellcolor[HTML]{1FA287}} \color[HTML]{F1F1F1} 0.62 & {\cellcolor[HTML]{24AA83}} \color[HTML]{F1F1F1} 0.64 & {\cellcolor[HTML]{433E85}} \color[HTML]{F1F1F1} 0.36 & {\cellcolor[HTML]{21A685}} \color[HTML]{F1F1F1} 0.63 & {\cellcolor[HTML]{2AB07F}} \color[HTML]{F1F1F1} 0.66 & {\cellcolor[HTML]{3BBB75}} \color[HTML]{F1F1F1} 0.69 & {\cellcolor[HTML]{3A538B}} \color[HTML]{F1F1F1} 0.41 & {\cellcolor[HTML]{21A685}} \color[HTML]{F1F1F1} 0.63 & {\cellcolor[HTML]{2FB47C}} \color[HTML]{F1F1F1} 0.67 & {\cellcolor[HTML]{42BE71}} \color[HTML]{F1F1F1} 0.7  \\
\textbf{R5}       & {\cellcolor[HTML]{481769}} \color[HTML]{F1F1F1} 0.28 & {\cellcolor[HTML]{1E9B8A}} \color[HTML]{F1F1F1} 0.6  & {\cellcolor[HTML]{1FA287}} \color[HTML]{F1F1F1} 0.62 & {\cellcolor[HTML]{27AD81}} \color[HTML]{F1F1F1} 0.65 & {\cellcolor[HTML]{414287}} \color[HTML]{F1F1F1} 0.37 & {\cellcolor[HTML]{21A685}} \color[HTML]{F1F1F1} 0.63 & {\cellcolor[HTML]{2FB47C}} \color[HTML]{F1F1F1} 0.67 & {\cellcolor[HTML]{42BE71}} \color[HTML]{F1F1F1} 0.7  & {\cellcolor[HTML]{38588C}} \color[HTML]{F1F1F1} 0.42 & {\cellcolor[HTML]{24AA83}} \color[HTML]{F1F1F1} 0.64 & {\cellcolor[HTML]{3BBB75}} \color[HTML]{F1F1F1} 0.69 & {\cellcolor[HTML]{52C569}} \color[HTML]{000000} 0.72 \\
\textbf{R6}       & {\cellcolor[HTML]{440154}} \color[HTML]{F1F1F1} 0.24 & {\cellcolor[HTML]{1F978B}} \color[HTML]{F1F1F1} 0.59 & {\cellcolor[HTML]{1FA287}} \color[HTML]{F1F1F1} 0.62 & {\cellcolor[HTML]{24AA83}} \color[HTML]{F1F1F1} 0.64 & {\cellcolor[HTML]{463480}} \color[HTML]{F1F1F1} 0.34 & {\cellcolor[HTML]{24AA83}} \color[HTML]{F1F1F1} 0.64 & {\cellcolor[HTML]{2FB47C}} \color[HTML]{F1F1F1} 0.67 & {\cellcolor[HTML]{42BE71}} \color[HTML]{F1F1F1} 0.7  & {\cellcolor[HTML]{38588C}} \color[HTML]{F1F1F1} 0.42 & {\cellcolor[HTML]{27AD81}} \color[HTML]{F1F1F1} 0.65 & {\cellcolor[HTML]{3BBB75}} \color[HTML]{F1F1F1} 0.69 & {\cellcolor[HTML]{4AC16D}} \color[HTML]{000000} 0.71 \\
\textbf{R7}       & {\cellcolor[HTML]{471164}} \color[HTML]{F1F1F1} 0.27 & {\cellcolor[HTML]{1E9B8A}} \color[HTML]{F1F1F1} 0.6  & {\cellcolor[HTML]{21A685}} \color[HTML]{F1F1F1} 0.63 & {\cellcolor[HTML]{27AD81}} \color[HTML]{F1F1F1} 0.65 & {\cellcolor[HTML]{3F4788}} \color[HTML]{F1F1F1} 0.38 & {\cellcolor[HTML]{24AA83}} \color[HTML]{F1F1F1} 0.64 & {\cellcolor[HTML]{35B779}} \color[HTML]{F1F1F1} 0.68 & {\cellcolor[HTML]{4AC16D}} \color[HTML]{000000} 0.71 & {\cellcolor[HTML]{365C8D}} \color[HTML]{F1F1F1} 0.43 & {\cellcolor[HTML]{27AD81}} \color[HTML]{F1F1F1} 0.65 & {\cellcolor[HTML]{3BBB75}} \color[HTML]{F1F1F1} 0.69 & {\cellcolor[HTML]{52C569}} \color[HTML]{000000} 0.72 \\
\textbf{R8}       & {\cellcolor[HTML]{450559}} \color[HTML]{F1F1F1} 0.25 & {\cellcolor[HTML]{20938C}} \color[HTML]{F1F1F1} 0.58 & {\cellcolor[HTML]{1FA287}} \color[HTML]{F1F1F1} 0.62 & {\cellcolor[HTML]{24AA83}} \color[HTML]{F1F1F1} 0.64 & {\cellcolor[HTML]{433E85}} \color[HTML]{F1F1F1} 0.36 & {\cellcolor[HTML]{1F9F88}} \color[HTML]{F1F1F1} 0.61 & {\cellcolor[HTML]{2AB07F}} \color[HTML]{F1F1F1} 0.66 & {\cellcolor[HTML]{3BBB75}} \color[HTML]{F1F1F1} 0.69 & {\cellcolor[HTML]{3A538B}} \color[HTML]{F1F1F1} 0.41 & {\cellcolor[HTML]{1FA287}} \color[HTML]{F1F1F1} 0.62 & {\cellcolor[HTML]{2FB47C}} \color[HTML]{F1F1F1} 0.67 & {\cellcolor[HTML]{4AC16D}} \color[HTML]{000000} 0.71 \\
\textbf{R9}       & {\cellcolor[HTML]{481769}} \color[HTML]{F1F1F1} 0.28 & {\cellcolor[HTML]{1E9B8A}} \color[HTML]{F1F1F1} 0.6  & {\cellcolor[HTML]{21A685}} \color[HTML]{F1F1F1} 0.63 & {\cellcolor[HTML]{2AB07F}} \color[HTML]{F1F1F1} 0.66 & {\cellcolor[HTML]{3C508B}} \color[HTML]{F1F1F1} 0.4  & {\cellcolor[HTML]{27AD81}} \color[HTML]{F1F1F1} 0.65 & {\cellcolor[HTML]{3BBB75}} \color[HTML]{F1F1F1} 0.69 & {\cellcolor[HTML]{52C569}} \color[HTML]{000000} 0.72 & {\cellcolor[HTML]{32648E}} \color[HTML]{F1F1F1} 0.45 & {\cellcolor[HTML]{2AB07F}} \color[HTML]{F1F1F1} 0.66 & {\cellcolor[HTML]{42BE71}} \color[HTML]{F1F1F1} 0.7  & {\cellcolor[HTML]{5AC864}} \color[HTML]{000000} 0.73 \\
\textbf{R10}      & {\cellcolor[HTML]{481769}} \color[HTML]{F1F1F1} 0.28 & {\cellcolor[HTML]{20938C}} \color[HTML]{F1F1F1} 0.58 & {\cellcolor[HTML]{1F9F88}} \color[HTML]{F1F1F1} 0.61 & {\cellcolor[HTML]{21A685}} \color[HTML]{F1F1F1} 0.63 & {\cellcolor[HTML]{414287}} \color[HTML]{F1F1F1} 0.37 & {\cellcolor[HTML]{1F9F88}} \color[HTML]{F1F1F1} 0.61 & {\cellcolor[HTML]{2AB07F}} \color[HTML]{F1F1F1} 0.66 & {\cellcolor[HTML]{3BBB75}} \color[HTML]{F1F1F1} 0.69 & {\cellcolor[HTML]{365C8D}} \color[HTML]{F1F1F1} 0.43 & {\cellcolor[HTML]{1FA287}} \color[HTML]{F1F1F1} 0.62 & {\cellcolor[HTML]{2FB47C}} \color[HTML]{F1F1F1} 0.67 & {\cellcolor[HTML]{42BE71}} \color[HTML]{F1F1F1} 0.7 \\
\bottomrule
\end{tabular}
\end{table*}

\begin{table*}
\centering
\caption{Results from the matching tasks for $\delta_{location}=30$ with linear weighting of textual similarity and spatial weight. Linear weighting achieved best results amongst $\delta_{location} = \{10,20,30\}$ at 30.}
\label{matching-results-ts}
\begin{tabular}{lrrrrrrrrr}
\toprule
Accuracy $\rightarrow$ & \multicolumn{3}{c}{\textbf{Top-1 (TS)}} & \multicolumn{3}{c}{\textbf{Top-2 (TS)}} & \multicolumn{3}{c}{\textbf{Top-3 (TS)}} \\
 \midrule
Lang $\downarrow$ K $\rightarrow$ & 25 & 50 & 100 & 25 & 50 & 100 & 25 & 50 & 100 \\
\midrule
\textbf{en} & {\cellcolor[HTML]{23898E}} \color[HTML]{F1F1F1} 0.58 & {\cellcolor[HTML]{23898E}} \color[HTML]{F1F1F1} 0.58 & {\cellcolor[HTML]{23898E}} \color[HTML]{F1F1F1} 0.58 & {\cellcolor[HTML]{90D743}} \color[HTML]{000000} 0.71 & {\cellcolor[HTML]{A2DA37}} \color[HTML]{000000} 0.72 & {\cellcolor[HTML]{A2DA37}} \color[HTML]{000000} 0.72 & {\cellcolor[HTML]{ECE51B}} \color[HTML]{000000} 0.76 & {\cellcolor[HTML]{FDE725}} \color[HTML]{000000} 0.77 & {\cellcolor[HTML]{FDE725}} \color[HTML]{000000} 0.77 \\
\midrule
\textbf{zh-CN} & {\cellcolor[HTML]{34618D}} \color[HTML]{F1F1F1} 0.52 & {\cellcolor[HTML]{34618D}} \color[HTML]{F1F1F1} 0.52 & {\cellcolor[HTML]{34618D}} \color[HTML]{F1F1F1} 0.52 & {\cellcolor[HTML]{2CB17E}} \color[HTML]{F1F1F1} 0.64 & {\cellcolor[HTML]{35B779}} \color[HTML]{F1F1F1} 0.65 & {\cellcolor[HTML]{40BD72}} \color[HTML]{F1F1F1} 0.66 & {\cellcolor[HTML]{90D743}} \color[HTML]{000000} 0.71 & {\cellcolor[HTML]{A2DA37}} \color[HTML]{000000} 0.72 & {\cellcolor[HTML]{B5DE2B}} \color[HTML]{000000} 0.73 \\
\textbf{zh-TW} & {\cellcolor[HTML]{3B528B}} \color[HTML]{F1F1F1} 0.5 & {\cellcolor[HTML]{375A8C}} \color[HTML]{F1F1F1} 0.51 & {\cellcolor[HTML]{34618D}} \color[HTML]{F1F1F1} 0.52 & {\cellcolor[HTML]{35B779}} \color[HTML]{F1F1F1} 0.65 & {\cellcolor[HTML]{40BD72}} \color[HTML]{F1F1F1} 0.66 & {\cellcolor[HTML]{4EC36B}} \color[HTML]{000000} 0.67 & {\cellcolor[HTML]{7FD34E}} \color[HTML]{000000} 0.7 & {\cellcolor[HTML]{A2DA37}} \color[HTML]{000000} 0.72 & {\cellcolor[HTML]{B5DE2B}} \color[HTML]{000000} 0.73 \\
\textbf{ar} & {\cellcolor[HTML]{3E4989}} \color[HTML]{F1F1F1} 0.49 & {\cellcolor[HTML]{3E4989}} \color[HTML]{F1F1F1} 0.49 & {\cellcolor[HTML]{3B528B}} \color[HTML]{F1F1F1} 0.5 & {\cellcolor[HTML]{20A486}} \color[HTML]{F1F1F1} 0.62 & {\cellcolor[HTML]{25AB82}} \color[HTML]{F1F1F1} 0.63 & {\cellcolor[HTML]{2CB17E}} \color[HTML]{F1F1F1} 0.64 & {\cellcolor[HTML]{5EC962}} \color[HTML]{000000} 0.68 & {\cellcolor[HTML]{6ECE58}} \color[HTML]{000000} 0.69 & {\cellcolor[HTML]{7FD34E}} \color[HTML]{000000} 0.7 \\
\textbf{vi} & {\cellcolor[HTML]{3B528B}} \color[HTML]{F1F1F1} 0.5 & {\cellcolor[HTML]{3B528B}} \color[HTML]{F1F1F1} 0.5 & {\cellcolor[HTML]{3B528B}} \color[HTML]{F1F1F1} 0.5 & {\cellcolor[HTML]{20A486}} \color[HTML]{F1F1F1} 0.62 & {\cellcolor[HTML]{25AB82}} \color[HTML]{F1F1F1} 0.63 & {\cellcolor[HTML]{25AB82}} \color[HTML]{F1F1F1} 0.63 & {\cellcolor[HTML]{5EC962}} \color[HTML]{000000} 0.68 & {\cellcolor[HTML]{6ECE58}} \color[HTML]{000000} 0.69 & {\cellcolor[HTML]{6ECE58}} \color[HTML]{000000} 0.69 \\
\textbf{el} & {\cellcolor[HTML]{34618D}} \color[HTML]{F1F1F1} 0.52 & {\cellcolor[HTML]{31688E}} \color[HTML]{F1F1F1} 0.53 & {\cellcolor[HTML]{31688E}} \color[HTML]{F1F1F1} 0.53 & {\cellcolor[HTML]{40BD72}} \color[HTML]{F1F1F1} 0.66 & {\cellcolor[HTML]{4EC36B}} \color[HTML]{000000} 0.67 & {\cellcolor[HTML]{5EC962}} \color[HTML]{000000} 0.68 & {\cellcolor[HTML]{A2DA37}} \color[HTML]{000000} 0.72 & {\cellcolor[HTML]{B5DE2B}} \color[HTML]{000000} 0.73 & {\cellcolor[HTML]{C8E020}} \color[HTML]{000000} 0.74 \\
\textbf{it} & {\cellcolor[HTML]{375A8C}} \color[HTML]{F1F1F1} 0.51 & {\cellcolor[HTML]{375A8C}} \color[HTML]{F1F1F1} 0.51 & {\cellcolor[HTML]{375A8C}} \color[HTML]{F1F1F1} 0.51 & {\cellcolor[HTML]{25AB82}} \color[HTML]{F1F1F1} 0.63 & {\cellcolor[HTML]{25AB82}} \color[HTML]{F1F1F1} 0.63 & {\cellcolor[HTML]{2CB17E}} \color[HTML]{F1F1F1} 0.64 & {\cellcolor[HTML]{6ECE58}} \color[HTML]{000000} 0.69 & {\cellcolor[HTML]{7FD34E}} \color[HTML]{000000} 0.7 & {\cellcolor[HTML]{90D743}} \color[HTML]{000000} 0.71 \\
\textbf{hi} & {\cellcolor[HTML]{481668}} \color[HTML]{F1F1F1} 0.43 & {\cellcolor[HTML]{481F70}} \color[HTML]{F1F1F1} 0.44 & {\cellcolor[HTML]{482878}} \color[HTML]{F1F1F1} 0.45 & {\cellcolor[HTML]{287C8E}} \color[HTML]{F1F1F1} 0.56 & {\cellcolor[HTML]{26828E}} \color[HTML]{F1F1F1} 0.57 & {\cellcolor[HTML]{23898E}} \color[HTML]{F1F1F1} 0.58 & {\cellcolor[HTML]{20A486}} \color[HTML]{F1F1F1} 0.62 & {\cellcolor[HTML]{25AB82}} \color[HTML]{F1F1F1} 0.63 & {\cellcolor[HTML]{35B779}} \color[HTML]{F1F1F1} 0.65 \\
\textbf{es} & {\cellcolor[HTML]{34618D}} \color[HTML]{F1F1F1} 0.52 & {\cellcolor[HTML]{34618D}} \color[HTML]{F1F1F1} 0.52 & {\cellcolor[HTML]{34618D}} \color[HTML]{F1F1F1} 0.52 & {\cellcolor[HTML]{35B779}} \color[HTML]{F1F1F1} 0.65 & {\cellcolor[HTML]{40BD72}} \color[HTML]{F1F1F1} 0.66 & {\cellcolor[HTML]{40BD72}} \color[HTML]{F1F1F1} 0.66 & {\cellcolor[HTML]{90D743}} \color[HTML]{000000} 0.71 & {\cellcolor[HTML]{A2DA37}} \color[HTML]{000000} 0.72 & {\cellcolor[HTML]{B5DE2B}} \color[HTML]{000000} 0.73 \\
\textbf{ko} & {\cellcolor[HTML]{3E4989}} \color[HTML]{F1F1F1} 0.49 & {\cellcolor[HTML]{3B528B}} \color[HTML]{F1F1F1} 0.5 & {\cellcolor[HTML]{3B528B}} \color[HTML]{F1F1F1} 0.5 & {\cellcolor[HTML]{20A486}} \color[HTML]{F1F1F1} 0.62 & {\cellcolor[HTML]{25AB82}} \color[HTML]{F1F1F1} 0.63 & {\cellcolor[HTML]{25AB82}} \color[HTML]{F1F1F1} 0.63 & {\cellcolor[HTML]{6ECE58}} \color[HTML]{000000} 0.69 & {\cellcolor[HTML]{90D743}} \color[HTML]{000000} 0.71 & {\cellcolor[HTML]{90D743}} \color[HTML]{000000} 0.71 \\
\textbf{gu} & {\cellcolor[HTML]{482878}} \color[HTML]{F1F1F1} 0.45 & {\cellcolor[HTML]{46307E}} \color[HTML]{F1F1F1} 0.46 & {\cellcolor[HTML]{46307E}} \color[HTML]{F1F1F1} 0.46 & {\cellcolor[HTML]{21908D}} \color[HTML]{F1F1F1} 0.59 & {\cellcolor[HTML]{1F978B}} \color[HTML]{F1F1F1} 0.6 & {\cellcolor[HTML]{1F978B}} \color[HTML]{F1F1F1} 0.6 & {\cellcolor[HTML]{35B779}} \color[HTML]{F1F1F1} 0.65 & {\cellcolor[HTML]{40BD72}} \color[HTML]{F1F1F1} 0.66 & {\cellcolor[HTML]{4EC36B}} \color[HTML]{000000} 0.67 \\
\textbf{id} & {\cellcolor[HTML]{375A8C}} \color[HTML]{F1F1F1} 0.51 & {\cellcolor[HTML]{375A8C}} \color[HTML]{F1F1F1} 0.51 & {\cellcolor[HTML]{34618D}} \color[HTML]{F1F1F1} 0.52 & {\cellcolor[HTML]{25AB82}} \color[HTML]{F1F1F1} 0.63 & {\cellcolor[HTML]{2CB17E}} \color[HTML]{F1F1F1} 0.64 & {\cellcolor[HTML]{35B779}} \color[HTML]{F1F1F1} 0.65 & {\cellcolor[HTML]{6ECE58}} \color[HTML]{000000} 0.69 & {\cellcolor[HTML]{90D743}} \color[HTML]{000000} 0.71 & {\cellcolor[HTML]{90D743}} \color[HTML]{000000} 0.71 \\
\textbf{fa} & {\cellcolor[HTML]{3E4989}} \color[HTML]{F1F1F1} 0.49 & {\cellcolor[HTML]{3B528B}} \color[HTML]{F1F1F1} 0.5 & {\cellcolor[HTML]{3B528B}} \color[HTML]{F1F1F1} 0.5 & {\cellcolor[HTML]{20A486}} \color[HTML]{F1F1F1} 0.62 & {\cellcolor[HTML]{2CB17E}} \color[HTML]{F1F1F1} 0.64 & {\cellcolor[HTML]{35B779}} \color[HTML]{F1F1F1} 0.65 & {\cellcolor[HTML]{4EC36B}} \color[HTML]{000000} 0.67 & {\cellcolor[HTML]{6ECE58}} \color[HTML]{000000} 0.69 & {\cellcolor[HTML]{6ECE58}} \color[HTML]{000000} 0.69 \\
\textbf{fr} & {\cellcolor[HTML]{3B528B}} \color[HTML]{F1F1F1} 0.5 & {\cellcolor[HTML]{3B528B}} \color[HTML]{F1F1F1} 0.5 & {\cellcolor[HTML]{3B528B}} \color[HTML]{F1F1F1} 0.5 & {\cellcolor[HTML]{25AB82}} \color[HTML]{F1F1F1} 0.63 & {\cellcolor[HTML]{2CB17E}} \color[HTML]{F1F1F1} 0.64 & {\cellcolor[HTML]{35B779}} \color[HTML]{F1F1F1} 0.65 & {\cellcolor[HTML]{6ECE58}} \color[HTML]{000000} 0.69 & {\cellcolor[HTML]{7FD34E}} \color[HTML]{000000} 0.7 & {\cellcolor[HTML]{90D743}} \color[HTML]{000000} 0.71 \\
\textbf{de} & {\cellcolor[HTML]{3B528B}} \color[HTML]{F1F1F1} 0.5 & {\cellcolor[HTML]{3B528B}} \color[HTML]{F1F1F1} 0.5 & {\cellcolor[HTML]{375A8C}} \color[HTML]{F1F1F1} 0.51 & {\cellcolor[HTML]{2CB17E}} \color[HTML]{F1F1F1} 0.64 & {\cellcolor[HTML]{35B779}} \color[HTML]{F1F1F1} 0.65 & {\cellcolor[HTML]{35B779}} \color[HTML]{F1F1F1} 0.65 & {\cellcolor[HTML]{7FD34E}} \color[HTML]{000000} 0.7 & {\cellcolor[HTML]{90D743}} \color[HTML]{000000} 0.71 & {\cellcolor[HTML]{A2DA37}} \color[HTML]{000000} 0.72 \\
\textbf{pt} & {\cellcolor[HTML]{31688E}} \color[HTML]{F1F1F1} 0.53 & {\cellcolor[HTML]{2E6F8E}} \color[HTML]{F1F1F1} 0.54 & {\cellcolor[HTML]{2E6F8E}} \color[HTML]{F1F1F1} 0.54 & {\cellcolor[HTML]{40BD72}} \color[HTML]{F1F1F1} 0.66 & {\cellcolor[HTML]{40BD72}} \color[HTML]{F1F1F1} 0.66 & {\cellcolor[HTML]{4EC36B}} \color[HTML]{000000} 0.67 & {\cellcolor[HTML]{A2DA37}} \color[HTML]{000000} 0.72 & {\cellcolor[HTML]{C8E020}} \color[HTML]{000000} 0.74 & {\cellcolor[HTML]{C8E020}} \color[HTML]{000000} 0.74 \\
\midrule
\textbf{R1} & {\cellcolor[HTML]{481F70}} \color[HTML]{F1F1F1} 0.44 & {\cellcolor[HTML]{481F70}} \color[HTML]{F1F1F1} 0.44 & {\cellcolor[HTML]{482878}} \color[HTML]{F1F1F1} 0.45 & {\cellcolor[HTML]{2B758E}} \color[HTML]{F1F1F1} 0.55 & {\cellcolor[HTML]{287C8E}} \color[HTML]{F1F1F1} 0.56 & {\cellcolor[HTML]{26828E}} \color[HTML]{F1F1F1} 0.57 & {\cellcolor[HTML]{1F9E89}} \color[HTML]{F1F1F1} 0.61 & {\cellcolor[HTML]{25AB82}} \color[HTML]{F1F1F1} 0.63 & {\cellcolor[HTML]{2CB17E}} \color[HTML]{F1F1F1} 0.64 \\
\textbf{R2} & {\cellcolor[HTML]{481668}} \color[HTML]{F1F1F1} 0.43 & {\cellcolor[HTML]{481F70}} \color[HTML]{F1F1F1} 0.44 & {\cellcolor[HTML]{481F70}} \color[HTML]{F1F1F1} 0.44 & {\cellcolor[HTML]{2E6F8E}} \color[HTML]{F1F1F1} 0.54 & {\cellcolor[HTML]{2B758E}} \color[HTML]{F1F1F1} 0.55 & {\cellcolor[HTML]{287C8E}} \color[HTML]{F1F1F1} 0.56 & {\cellcolor[HTML]{1F978B}} \color[HTML]{F1F1F1} 0.6 & {\cellcolor[HTML]{1F9E89}} \color[HTML]{F1F1F1} 0.61 & {\cellcolor[HTML]{25AB82}} \color[HTML]{F1F1F1} 0.63 \\
\textbf{R3} & {\cellcolor[HTML]{481668}} \color[HTML]{F1F1F1} 0.43 & {\cellcolor[HTML]{481668}} \color[HTML]{F1F1F1} 0.43 & {\cellcolor[HTML]{481F70}} \color[HTML]{F1F1F1} 0.44 & {\cellcolor[HTML]{2E6F8E}} \color[HTML]{F1F1F1} 0.54 & {\cellcolor[HTML]{2B758E}} \color[HTML]{F1F1F1} 0.55 & {\cellcolor[HTML]{287C8E}} \color[HTML]{F1F1F1} 0.56 & {\cellcolor[HTML]{1F978B}} \color[HTML]{F1F1F1} 0.6 & {\cellcolor[HTML]{1F9E89}} \color[HTML]{F1F1F1} 0.61 & {\cellcolor[HTML]{20A486}} \color[HTML]{F1F1F1} 0.62 \\
\textbf{R4} & {\cellcolor[HTML]{460B5E}} \color[HTML]{F1F1F1} 0.42 & {\cellcolor[HTML]{481668}} \color[HTML]{F1F1F1} 0.43 & {\cellcolor[HTML]{481F70}} \color[HTML]{F1F1F1} 0.44 & {\cellcolor[HTML]{2E6F8E}} \color[HTML]{F1F1F1} 0.54 & {\cellcolor[HTML]{2B758E}} \color[HTML]{F1F1F1} 0.55 & {\cellcolor[HTML]{287C8E}} \color[HTML]{F1F1F1} 0.56 & {\cellcolor[HTML]{21908D}} \color[HTML]{F1F1F1} 0.59 & {\cellcolor[HTML]{1F9E89}} \color[HTML]{F1F1F1} 0.61 & {\cellcolor[HTML]{20A486}} \color[HTML]{F1F1F1} 0.62 \\
\textbf{R5} & {\cellcolor[HTML]{460B5E}} \color[HTML]{F1F1F1} 0.42 & {\cellcolor[HTML]{481668}} \color[HTML]{F1F1F1} 0.43 & {\cellcolor[HTML]{481668}} \color[HTML]{F1F1F1} 0.43 & {\cellcolor[HTML]{31688E}} \color[HTML]{F1F1F1} 0.53 & {\cellcolor[HTML]{2E6F8E}} \color[HTML]{F1F1F1} 0.54 & {\cellcolor[HTML]{2B758E}} \color[HTML]{F1F1F1} 0.55 & {\cellcolor[HTML]{1F978B}} \color[HTML]{F1F1F1} 0.6 & {\cellcolor[HTML]{1F9E89}} \color[HTML]{F1F1F1} 0.61 & {\cellcolor[HTML]{20A486}} \color[HTML]{F1F1F1} 0.62 \\
\textbf{R6} & {\cellcolor[HTML]{481F70}} \color[HTML]{F1F1F1} 0.44 & {\cellcolor[HTML]{482878}} \color[HTML]{F1F1F1} 0.45 & {\cellcolor[HTML]{482878}} \color[HTML]{F1F1F1} 0.45 & {\cellcolor[HTML]{287C8E}} \color[HTML]{F1F1F1} 0.56 & {\cellcolor[HTML]{26828E}} \color[HTML]{F1F1F1} 0.57 & {\cellcolor[HTML]{23898E}} \color[HTML]{F1F1F1} 0.58 & {\cellcolor[HTML]{1F9E89}} \color[HTML]{F1F1F1} 0.61 & {\cellcolor[HTML]{20A486}} \color[HTML]{F1F1F1} 0.62 & {\cellcolor[HTML]{25AB82}} \color[HTML]{F1F1F1} 0.63 \\
\textbf{R7} & {\cellcolor[HTML]{481668}} \color[HTML]{F1F1F1} 0.43 & {\cellcolor[HTML]{481668}} \color[HTML]{F1F1F1} 0.43 & {\cellcolor[HTML]{481F70}} \color[HTML]{F1F1F1} 0.44 & {\cellcolor[HTML]{2E6F8E}} \color[HTML]{F1F1F1} 0.54 & {\cellcolor[HTML]{2B758E}} \color[HTML]{F1F1F1} 0.55 & {\cellcolor[HTML]{287C8E}} \color[HTML]{F1F1F1} 0.56 & {\cellcolor[HTML]{1F9E89}} \color[HTML]{F1F1F1} 0.61 & {\cellcolor[HTML]{25AB82}} \color[HTML]{F1F1F1} 0.63 & {\cellcolor[HTML]{25AB82}} \color[HTML]{F1F1F1} 0.63 \\
\textbf{R8} & {\cellcolor[HTML]{440154}} \color[HTML]{F1F1F1} 0.41 & {\cellcolor[HTML]{440154}} \color[HTML]{F1F1F1} 0.41 & {\cellcolor[HTML]{460B5E}} \color[HTML]{F1F1F1} 0.42 & {\cellcolor[HTML]{31688E}} \color[HTML]{F1F1F1} 0.53 & {\cellcolor[HTML]{2E6F8E}} \color[HTML]{F1F1F1} 0.54 & {\cellcolor[HTML]{2B758E}} \color[HTML]{F1F1F1} 0.55 & {\cellcolor[HTML]{23898E}} \color[HTML]{F1F1F1} 0.58 & {\cellcolor[HTML]{21908D}} \color[HTML]{F1F1F1} 0.59 & {\cellcolor[HTML]{1F9E89}} \color[HTML]{F1F1F1} 0.61 \\
\textbf{R9} & {\cellcolor[HTML]{481668}} \color[HTML]{F1F1F1} 0.43 & {\cellcolor[HTML]{481F70}} \color[HTML]{F1F1F1} 0.44 & {\cellcolor[HTML]{482878}} \color[HTML]{F1F1F1} 0.45 & {\cellcolor[HTML]{2B758E}} \color[HTML]{F1F1F1} 0.55 & {\cellcolor[HTML]{287C8E}} \color[HTML]{F1F1F1} 0.56 & {\cellcolor[HTML]{26828E}} \color[HTML]{F1F1F1} 0.57 & {\cellcolor[HTML]{20A486}} \color[HTML]{F1F1F1} 0.62 & {\cellcolor[HTML]{25AB82}} \color[HTML]{F1F1F1} 0.63 & {\cellcolor[HTML]{2CB17E}} \color[HTML]{F1F1F1} 0.64 \\
\textbf{R10} & {\cellcolor[HTML]{481F70}} \color[HTML]{F1F1F1} 0.44 & {\cellcolor[HTML]{481F70}} \color[HTML]{F1F1F1} 0.44 & {\cellcolor[HTML]{482878}} \color[HTML]{F1F1F1} 0.45 & {\cellcolor[HTML]{31688E}} \color[HTML]{F1F1F1} 0.53 & {\cellcolor[HTML]{2E6F8E}} \color[HTML]{F1F1F1} 0.54 & {\cellcolor[HTML]{2B758E}} \color[HTML]{F1F1F1} 0.55 & {\cellcolor[HTML]{21908D}} \color[HTML]{F1F1F1} 0.59 & {\cellcolor[HTML]{1F978B}} \color[HTML]{F1F1F1} 0.6 & {\cellcolor[HTML]{1F978B}} \color[HTML]{F1F1F1} 0.6 \\
\bottomrule
\end{tabular}
\end{table*}

Results show that incorporating textual, temporal, and spatial features collectively yields improved performance. We observed that an increase in $K$ leads to higher matching accuracy. This observation holds true for our approach across Top-$n$ conditions. The reason behind this is that a larger value of $K$ results in a greater number of matches being retrieved through semantic search. However, for ``T'', the accuracy remains consistent across the same Top-$n$ results but increases when considering different values of $K$. This is because our approach utilizes temporal and spatial weights to re-evaluate and prioritize the matched results, introducing improved accuracy within the same Top-$n$ results as well. ``TS'' also improves with increase in $K$ within a Top-$n$, however, the improvement is minimal compared to ``TTS''.

\begin{table*}
    \centering
    \caption{Some examples from the matching tasks (language=$en$, $k$=100, $\delta_{time} =30$ and $\delta_{location} =10$) are provided below. The best-matched offer by the respective method is included.}
    \label{examples_of_match}
    \begin{tabular}{p{4cm}|p{4cm}|p{4cm}|p{4cm}}
    \toprule
    \textbf{Request} & \textbf{Offer (TTS)} & \textbf{Offer (TS)} & \textbf{Offer (T)}\\
    \midrule
    North Brisbane is in dire need of emergency food and water after the recent disaster. Please help us if you can! & To those in need in North Brisbane, I'm prepared to provide emergency food and water. Reach out, and I'll do my best to assist! \newline \texttt{\{distance=0.3kms, time=0.06days\}} & I'm prepared to provide emergency food and water assistance to those in North Brisbane. If you or anyone in the area needs help, please reach out. We're in this together! \newline \texttt{\{distance=0.43kms, time=101.9days\}} & Offering emergency food and water assistance to those affected by the crisis in Melbourne. Your well-being is important. Reach out for support! \newline \texttt{\{distance=1375.54kms, time=331.8days\}} \\
    \midrule
    Urgent appeal in Melbourne: Gas leak in the area, and many families are evacuated without warm clothing and blankets. Please help us stay warm! & Just got word about the gas leak. I've got warm clothing and blankets ready. Let's support those affected! \newline \texttt{\{distance=0.55kms, time=0.11days\}} & Ready to assist with evacuations due to a gas leak in North Melbourne. If you're in the affected area, reach out for help. Stay safe! \newline \texttt{\{distance=1.29kms, time=461days\}} & Ready to assist with evacuations due to a gas leak in North Melbourne. If you're in the affected area, reach out for help. Stay safe! \newline \texttt{\{distance=1.29kms, time=461days\}} \\
    \midrule
      Urgently need a quarantine facility in Adelaide for a family member showing symptoms. Please help! &  I'm ready to assist Adelaide residents with quarantine facilities. Reach out if you need help! \newline \texttt{\{distance=0.9kms, time=0.05days\}} & Adelaide Emergency Services is coordinating quarantine facilities. Contact us for immediate assistance and support. \newline \texttt{\{distance=0.3kms, time=95.28days\}} & HealthCare Solutions is offering quarantine facilities in Adelaide. Connect with us for immediate assistance. \newline \texttt{\{distance=1.5kms, time=176days\}}\\
      \bottomrule
    \end{tabular}
\end{table*}

\begin{table*}
    \centering
    \caption{Some mismatched examples with ''TTS" method (language=$en$, $k$=100, $\delta_{time} =30$ and $\delta_{location} =10$).}
    \label{examples_of_mismatch}
    \begin{tabular}{p{4.4cm}|p{6cm}|p{6cm}}
    \toprule
    \textbf{Request} & \textbf{Offer (TTS)} & \textbf{True offer} \\
    \midrule
    Southeast Brisbane facing a mudslide threat. Need help reinforcing barriers and securing vulnerable areas. & Offering my help for cleanup and relief efforts in areas affected by mudslides in Brisbane. Let's join forces to restore our city. \#BrisbaneTogether \newline \texttt{\{distance=2.67kms, time=2.85days\}} & I have experience in mudslide prevention. Ready to assist in reinforcing barriers and securing areas at risk! \newline \texttt{\{distance=2.89kms, time=0.05days\}} \\
    \midrule
    Emergency evacuation due to a landslide in Adelaide. Need shelter urgently! & To those affected by the floods in Adelaide, I'm offering temporary shelter. Stay strong and reach out for assistance! \newline \texttt{\{distance=2.5kms, time=0.68days\}} & For those in Adelaide affected by the landslide, I can provide temporary shelter. Stay safe and reach out for support! \newline \texttt{\{distance=3.35kms, time=0.1days\}} \\
    \midrule
    In search of a quarantine facility in Melbourne for my family. COVID-19 has hit us hard, and we need a safe space to isolate. Please help! & I understand these times have been challenging for many. If anyone in Melbourne needs someone to talk to or needs support, I'm available.  \#SupportForAll \newline \texttt{\{not an exact offer\}} & I'm ready to provide a quarantine space for a family in need. Let's support each other during these challenging times. Reach out if you require help. \newline \texttt{\{distance=1.35kms, time=0.06days\}}\\
      \bottomrule
    \end{tabular}
\end{table*}

With our approach, the matching tasks exhibited the best performance with $k=100$, $\delta_{time} = 30$, and $\delta_{location} = 10$, amongst all experiments. Due to space constraints, we present results (in Table \ref{matching-results}) from $\delta_{time} =30$ and $\delta_{location} =10$ for all $K$, and make them a reference for discussions. Refer to Table \ref{examples_of_match} for matched examples and Table \ref{examples_of_mismatch} for mismatched examples. Between the lowest and highest values of $K$, both Top-2 and Top-3 showed higher improvements compared to Top-1. This is justified since achieving a perfect match in the Top-1 results is a challenging task, considering the possibility that an offer, perfectly suited for a particular request, might also be an ideal match for other requests (refer to first and second examples in Table \ref{examples_of_mismatch}) due to the random generation of time and location values. We also present results for ``TS'' in Table \ref{matching-results-ts}; we limit the results due to space constraints to $\delta_{location} =30$, where linear weighing achieved the best results. Except for Gujarati, ``TTS'' performed better across all languages and random datasets. Next, we discuss the results of our approach in detail.

We observed declining performances as the values of $\delta_{time}$ and $\delta_{location}$ increased. As these values increase, the number of potential offers increases, and finding the one offer hardcoded in the dataset becomes challenging. When $\delta_{distance} = 10$, we observed a decrease in accuracy over time: a 2\% decrease on average at $\delta_{time} = 90$ and a 4\% decrease on average at $\delta_{time} = 180$. Similarly, when $\delta_{distance} = 20$, we observed a 3\% decrease on average at $\delta_{time} = 90$ and a 5\% decrease on average at $\delta_{time} = 180$. $\delta_{distance} = 30$ showed similar trend as $\delta_{distance} = 20$.

The English dataset achieved the highest matching accuracy of 0.9, followed by Portuguese and Indonesian (0.88), Italian, German and Greek (0.87), Spanish and French (0.86), Vietnamese (0.85), Chinese Simplified (0.83), Korean (0.83), Chinese Traditional (0.82), Arabic (0.78), Farsi (0.72), Hindi (0.69), and Gujarati (0.61). The random datasets consistently showed accuracy in the 0.70--0.73 range. The higher accuracy for English and European languages can be attributed to the availability of more training data from those languages in the parallel datasets considered during the training of the sentence encoder \cite{lamsal2023crosslingual}. Within the random datasets, the lowest accuracy was observed in \textit{R4} and \textit{R10}, where the prevalence of the Gujarati and Hindi (respectively) corresponded to the lowest accuracy. This observation suggests the critical importance of the sentence encoder's ability to generate semantically meaningful sentence embeddings. This significance becomes evident when scrutinizing accuracies across textual features. Even when considering only textual features, the accuracy for the English language reaches as high as 0.64, and for European languages, it surpasses 0.54. However, for low-resource languages such as Gujarati and Hindi, the highest accuracies are 0.43 and 0.45, respectively.

\begin{table}
\centering
\caption{Statistics of textual similarities where offers corrected matched with ``TTS'' (Top-3 at K=100).}
\label{similarity-stats}
\begin{tabular}{p{1.5cm}|l|lll|l}
\toprule
\textbf{Dataset} & \textbf{Mean} & \textbf{Q1} & \textbf{Q2} & \textbf{Q3} & \textbf{Max.} \\ \midrule
\textbf{en}      & 0.43          & 0.31          & 0.43          & 0.55          & 0.84         \\ \midrule
\textbf{zh-CN}   & 0.39          & 0.27          & 0.39          & 0.52          & 0.80          \\
\textbf{zh-TW}   & 0.37          & 0.24          & 0.37          & 0.50          & 0.79         \\
\textbf{ar}      & 0.35          & 0.21          & 0.36          & 0.48          & 0.74         \\
\textbf{vi}      & 0.44          & 0.32          & 0.46          & 0.57          & 0.88         \\
\textbf{el}      & 0.41          & 0.29          & 0.42          & 0.54          & 0.84         \\
\textbf{it}      & 0.46          & 0.34          & 0.46          & 0.58          & 0.82         \\
\textbf{hi}      & 0.33          & 0.20           & 0.32          & 0.46          & 0.86         \\
\textbf{es}      & 0.44          & 0.33          & 0.45          & 0.57          & 0.82         \\
\textbf{ko}      & 0.38          & 0.26          & 0.37          & 0.50           & 0.83         \\
\textbf{gu}      & 0.27          & 0.13          & 0.26          & 0.37          & 0.80         \\
\textbf{id}      & 0.44          & 0.32          & 0.44          & 0.57          & 0.88         \\
\textbf{fa}      & 0.33          & 0.20           & 0.33          & 0.45          & 0.82         \\
\textbf{fr}      & 0.45          & 0.33          & 0.47          & 0.58          & 0.87         \\
\textbf{de}      & 0.45          & 0.35          & 0.47          & 0.57          & 0.82         \\
\textbf{pt}      & 0.44          & 0.33          & 0.46          & 0.57          & 0.85         \\ \bottomrule
\end{tabular}
\end{table}

We examined cases where our approach correctly matched offers in the Top-3 at K=100. For each successful match, we assessed textual similarity. Descriptive statistics of these similarities are presented in Table \ref{similarity-stats}, which includes the mean, quartiles (Q1, Q2, Q3), and maximum values. These statistics provide an overview of similarity scores across various datasets, which assists in understanding the typical range and variability of similarity scores for correctly matched offers using our approach. Results show higher mean similarity for languages such as English (0.43), Italian (0.46), Spanish (0.44), French (0.45), and German (0.45), while the lowest mean accuracies are observed in Hindi (0.33), Farsi (0.33), and Gujarati (0.27). These results are consistent with what we observed regarding matching accuracies across datasets, confirming that textual features play a critical role in the matching tasks as temporal and spatial weights are, in fact, weighing the textual similarity. These results can serve as a reference for future studies to set similarity thresholds across languages for similar matching tasks in crisis informatics.

\subsection{Index/search times and accuracy trade-offs}
We further examined the trade-offs between index/search times and accuracy of our approach, building upon the findings presented in Table \ref{matching-results}, specifically focusing on the English dataset and its Top-3 accuracy. Table \ref{time-accuracy} presents a comparison of different search strategies based on their index time, search time, and Top-3 accuracy for various values of $K$. The exhaustive search involves exploring the entire search space; it has the lowest index times (1.42ms on avg.) and highest accuracies (0.85--0.9), resulting in the highest search times (24.52--33.61ms) compared to other methods, which are approximation-based.

IVF with just two partitions ($part.$) and a single cell search, i.e., probe ($np$), reduced the search times to 9.99--10.81ms while compromising accuracy (0.72--0.75) and indexing times (60.94ms). Due to the dataset size, $part. > 3$ had higher search times even than exhaustive search. So, we limit to $part. = {1,2,3}$. More partitions and probes tend to improve accuracy. We observed the best performance with IVF at 3 partitions and 2 probes; accuracies (0.82--0.86) and search times (10.07--10.81ms). Next, we introduced PQ to IVF for improving search times. Considering the dataset size, we experimented with quantization centroids $m = {8,16,32}$ and bits for representing each vector after quantization as $4$. Introducing PQ improved search times (3.52--5.97ms) compared to IVF. IVF + PQ performed better at higher values of $K$ with comparable accuracies against IVF. Higher $np$ also increased the matching performance of IVF + PQ, making the composite strategy competitive against IVF along with reducing search times. On the other hand, the graph-based approach, HNSW, provided competitive matching accuracy compared to exhaustive search while maintaining both the index and search times lower. The indexing times with IVF and IVF + PQ were between 60.94--1715.42ms. HNSW took 15.29ms for indexing and 6.87--8.49ms for searching while matching the performance of the exhaustive search.

\begin{table*}
\caption{Comparison of search methods: index/search times (\textit{milliseconds}) and Top-3 accuracy trade-offs.}
\label{time-accuracy}
\centering
\begin{tabular}{lllrlrlr}
\toprule
&   & \multicolumn{2}{c}{\textbf{k=25}}  & \multicolumn{2}{c}{\textbf{k=50}}                       & \multicolumn{2}{c}{\textbf{k=100}}   \\
\midrule
& \multicolumn{1}{l}{\textbf{Index time}} & \multicolumn{1}{c}{\textbf{Search time}} & \multicolumn{1}{c}{\textbf{Acc.}} & \multicolumn{1}{c}{\textbf{Search time}} & \multicolumn{1}{c}{\textbf{Acc.}} & \multicolumn{1}{c}{\textbf{Search time}} & \multicolumn{1}{c}{\textbf{Acc.}} \\
\midrule
\textbf{Exhaustive search}                       & 0.53--8.2 (1.42)                       & 19.06--46.22 (24.52)                    & 0.85                                  & 16.71--41.70 (26.69)                    & 0.88                                  & 18.61--45.04 (33.61)                    & 0.9                                   \\
\midrule
\textbf{IVF} (part. = 2, np=1)               & \multirow{3}{*}{20.34--121.51 (60.94)} & 8.64--12.04 (9.99)                      & 0.72                                  & 9.29--11.84 (10.16)                     & 0.73                                  & 9.79--11.94 (10.81)                     & 0.75                                  \\
\textbf{IVF} (part. = 3, np=1)               &                                       & 5.88--6.46 (6.09)                       & 0.74                                  & 6.19--7.21 (6.51)                       & 0.76                                  & 6.64--7.34 (6.9)                        & 0.77                                  \\
\midrule
\textbf{IVF} (part. = 3, np=2)     &                                       & 9.4--11.05 (10.07)                      & 0.82                                  & 9.73--10.72 (10.22)                     & 0.84                                  & 10.31--11.43 (10.81)                    & 0.86                                  \\
\midrule
\multicolumn{8}{l}{part.=3, bits=4, np=1} \\
\textbf{IVF + PQ} (m=8)                    & 294.29--1524.95 (760.06)               & 2.90--3.6 (3.39)                        & 0.64                                  & 3.384--3.82 (3.52)                      & 0.73                                  & 3.43--9.79 (4.40)                       & 0.76                                  \\
\textbf{IVF + PQ} (m=16)                   & 381.11--1854.66 (1108.79)              & 3.164--3.73 (3.45)                      & 0.71                                  & 3.57--4.02 (3.76)                       & 0.75                                  & 3.72--4.7 (4)                           & 0.78                                  \\
\textbf{IVF + PQ} (m=32)                   & 927.15--2695.84 (1715.42)              & 3.63--4.05 (3.89)                       & 0.73                                  & 3.80--4.20 (4.06)                       & 0.77                                  & 4.31--4.99 (4.42)                       & 0.79                                  \\
\midrule
\multicolumn{8}{l}{part.=3, bits=4, np=2} \\
\textbf{IVF + PQ} (m=8)                   &                                       & 2.94--3.74 (3.51)                       & 0.68                                  & 3.34--6.93 (3.97)                       & 0.78                                  & 3.86--4.24 (4.08)                       & 0.83                                  \\
\textbf{IVF + PQ}  (m=16)                  &                                       & 3.86--4.35 (4.06)                       & 0.77                                  & 3.81--4.33 (4.07)                       & 0.81                                  & 4.38--4.79 (4.53)                       & 0.85                                  \\
\textbf{IVF + PQ} (m=32)                   &                                       & 4.68--5.23 (4.93)                       & 0.78                                  & 5.05--5.52 (5.304)                      & 0.83                                  & 5.69--6.6 (5.97)                        & 0.86                                  \\
\midrule
\textbf{HNSW} (s=16, c=16) & 11.88--25.76 (15.29)                   & 5.76--7.9 (6.87)                        & 0.84                                  & 6.45--9.11 (7.76)                       & 0.87                                  & 7.37--10.62 (8.49)                      & 0.89        \\
\bottomrule
\end{tabular}
\end{table*}

\subsection{Analysis of a million-scale real-world dataset}

\begin{table*}
    \centering
    \caption{A sample of tweets in MegaGeoCOV classified as requests and offers.}
    \label{classification-megageocov}
    \begin{tabular}{p{15cm}c}
    \toprule
    \textbf{Tweet} & \textbf{Classified as} \\
    \midrule
    Just shy of 5 weeks until I travel to Uganda to teach with some amazing health care workers. If your in a position to donate we would be really grateful. Paediatric Resuscitation Training, CHI Global in partnership with Nurture Africa, Uganda [HTTPURL] \#iDonate\_ie  & \multirow{11}{*}{request} \\
    \vspace{0.01cm}
    @MENTION Support us and help all the covid patients and daily wage workers. lockdown is not same for everyone, We are trying our level best with our existing resources and so can you. We are running a helpline number [phone] to help covid patients.\#donateyef @MENTION [HTTPURL]     &  \\
    \vspace{0.01cm}
    Join Us. Be a Volunteer. \#charity \#nonprofit \#donate \#love \#community \#covid \#fundraising \#support \#help \#volunteer \#giveback \#donation ... [HTTPURL] &  \\
    \vspace{0.01cm}
    Please help @MENTION feed the medical professionals at various hospitals who are on our front lines battling this invisible virus. The meals will start next week. Here's the… [HTTPURL] & \\
    \midrule
    Tomorrow, staring at 3pm, will hand out KN95 Face Masks, Hand Sanitizers, face shields and Take home COVID19 tests till supplies last.  \#stopcovid19  @MENTION @ Pelham Parkway, Bronx [HTTPURL] & \multirow{10}{*}{offer} \\
    \vspace{0.01cm}
    Please contact us if you need help in getting Covid vaccine. \#VaccinesWork \#worldimmunizationweek \#vaccine \#pharmacy \#pharmacist \#chemist @ TerryWhite Chemmart Cumberland Park Pharmacy [HTTPURL]. & \\
    \vspace{0.01cm}
    I have assembled a collective of psychiatrists, mental health workers, and clergy who are ready and willing to offer a safe and open space to listen to our traumatized students. &  \\
    \vspace{0.01cm}
    The [agency] will be donating marketing \& branding services to those businesses directly impacted by Covid-19. Visit [HTTPURL] and contact us today.  \#dontwaitcreate \#weareinthistogether & \\
    \bottomrule
    \end{tabular}

\end{table*}

The MegaGeoCOV Extended dataset \cite{lamsal2023narrative} comprised 33.9 million multi-lingual geo-tagged tweets. As the regular expressions and classifiers utilized for identifying requests and offers were tailored for English, non-English tweets were filtered out. Among the 17.8 million English tweets, 40\% (7.2 million tweets) matched at least one regular expression, indicating potential requests and offers. These 7.2 million tweets underwent classification to identify exclusive requests and offers. Among them, our classifiers identified 210k tweets seeking assistance and 72k tweets offering help. A summary of the findings is presented in Table \ref{megageocov-analysis}. Each classifier took less than 2 hours to classify 7.2 million tweets on an NVIDIA A100 GPU (80GB). Some request and offer tweets from the classification are listed in Table \ref{classification-megageocov}. We conducted a random sampling of 200 tweets categorized as requests and offers. Upon manual inspection, we observed that only 41\% of the tweets classified as requests were actually soliciting help during the COVID-19 crisis. Similarly, among offer tweets, only 38\% explicitly offered assistance. The remaining were more general requests and offers. Examples include:

\texttt{Among requests}:

\textit{Suicide figures are up 200\% since lockdown. Could 2 friends please copy and re-post this tweet? We’re trying to demonstrate that someone is always listening.}

\textit{Please wear a mask, please.}

\textit{STAY HOME, STAY SAFE AND SHOP ONLINE with us by clicking on the link in our bio or call [phone] to order and make further enquiries.}

\texttt{Among offers}:

\textit{While things are closing and shutting down due to the coronavirus, Reviewcade is still happening this week. We want to keep you guys entertained and help keep your minds off of what is going on and escape. We will be finishing Aladdin this week too!
}

\textit{I'm thinking about doing a quarantine series called "quarantine comfort food". I'm wanting to make tamales! Would that be something y'all wanna see?
}

\begin{table}
\centering
\caption{Classification tasks on \textit{MegaGeoCOV Extended} \cite{lamsal2023narrative}.}
\label{megageocov-analysis}
\begin{tabular}{lllll}
\toprule
\multicolumn{2}{l}{\multirow{2}{*}{\textbf{Potential request/offer tweets}}} & \multirow{2}{*}{7,256,139} & \textbf{Requests} & 210,072            \\
\multicolumn{2}{l}{}           &                            & \textbf{Offers}   & 72,406             \\ 
\midrule
\textbf{Country}               & \textbf{All tweets}           & \textbf{En tweets}         & \textbf{Potential}      & \textbf{O:R} \\
\midrule
(1) United States              & 10.54m                        & 8.81m                      & 3.98m                   & 0.36               \\
(2) United Kingdom             & 3.25m                         & 2.84m                      & 1.05m                   & 0.42               \\
. & . & . & . & .
\\
(5) Australia                  & 0.56m                         & 0.48m                      & 0.18m                   & 0.53      \\
\bottomrule
\end{tabular}
\end{table}

\begin{figure*}
    \centering      
    \includegraphics[width=0.7\textwidth]{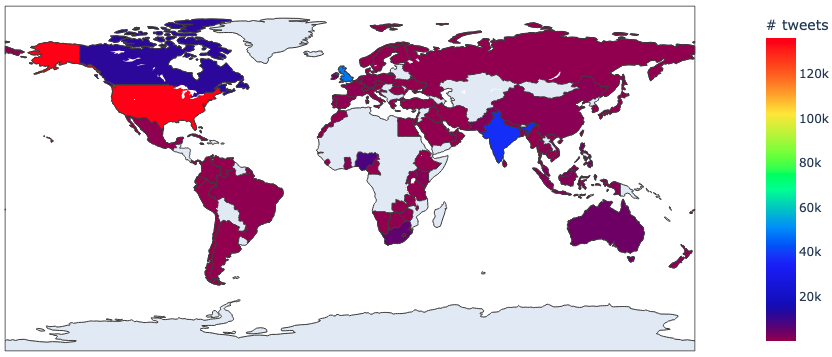}
    \caption{Global distribution of request and offer tweets during the COVID-19 pandemic. Countries considered: \# geotagged tweets $>$ 50.}
    \label{global-dist}
\end{figure*}

\begin{figure*}
    \centering
    \includegraphics[width=0.7\textwidth]{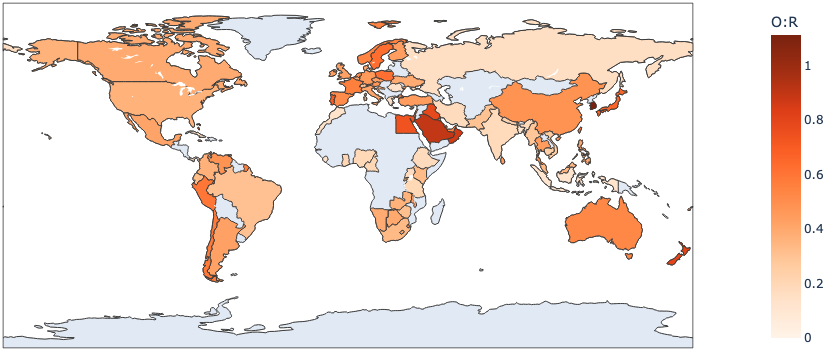}
    \caption{Offers to requests ratio ($O:R$) of countries considered in Figure \ref{global-dist}.}
    \label{or-global-dist}
\end{figure*}

We used the classification results to explore the distribution of crisis communications concerning requests and offer during the pandemic. The top contributors to the COVID-19 English discourse, included the United States, the United Kingdom, India, Canada, and Australia, collectively representing over 81\% of the discussion. Among the top 10 countries, Australia had the highest \textit{ratio of offers to requests} ($O:R$), at 0.53, indicating a potential difference in community engagement or resources available for assistance. Apart from the ones in Table \ref{megageocov-analysis}, Canada had $O:R$ of 0.39, India 0.18, South Africa 0.33, Nigeria 0.19, Ireland 0.43, the Philippines 0.39, and Pakistan 0.30. We observed regional variations in $O:R$ within a country, attributed to factors such as population density and contemporaneous issues (refer Table \ref{top-regions}). In the US, Los Angeles had $O:R$ of 0.30, Manhattan 0.16, Chicago 0.33, Houston 0.4. For UK regions, South East had 0.09, Manchester 0.4, Edinburgh 0.34, London 0.3, Glasgow 0.55, and Birmingham 0.51. Similarly, for Australian cities, Melbourne 0.61, Sydney and Brisbane 0.57. We present global distribution of request and offer tweets for the countries with $>50$ tweets in Figure \ref{global-dist}, and their respective $O:R$ in Figure \ref{or-global-dist}. Results from $O:R$ suggest that maintaining limited $\delta_{distance}$ can be problematic for some regions as the number of requests are much higher compared to offers. 

\begin{table}
\centering
\caption{Top active regions in the US, the UK and Australia.}
\label{top-regions}
\begin{tabular}{lll}
\toprule
\textbf{Country} & \textbf{Type} & \textbf{Regions (sorted (\# tweets) $\rightarrow$)}                     \\
\midrule
United States    & Request       & Los Angeles, Manhattan, Chicago          \\
                 & Offer         & Los Angeles,  Houston, Manhattan         \\
                 \midrule
United Kingdom   & Request       & South East, Manchester, Edinburgh \\
                 & Offer         & Glasgow,  Birmingham, Manchester         \\
                 \midrule
Australia        & Request       & Melbourne, Sydney, Brisbane              \\
                 & Offer         & Melbourne, Sydney, Brisbane             \\
\bottomrule
\end{tabular}
\end{table}

\section{Conclusion}
\label{conclusion}

In this study, we proposed a systematic approach to integrate textual, temporal, and spatial features for the identification and matching of requests and offers shared on social media. We trained classifiers to identify if a text is asking for assistance, offering assistance, or irrelevant. Results showed that CrisisTransformers, a family of crisis-domain-specific pre-trained models, outperform strong baselines like RoBERTa, MPNet, and BERTweet in classification tasks of identifying request and offer texts. Texts classified as requests and offers are then processed through a sentence encoder to generate sentence embeddings. We used CrisisTransformers' multi-lingual sentence encoder for this task. The encoder outperforms traditional embedding approaches such as word2vec and GloVe, as well as more sophisticated models such as Universal Sentence Encoder and Sentence Transformers. We show that a cross-lingual embedding space is effective in generating sentence embeddings required for the matching tasks where requests are matched with relevant offers. Next, we experimented with different vector search strategies and studied their effects on indexing/searching times and accuracy. Furthermore, we analyzed a million-scale geotagged tweets dataset to study crisis communications concerning requests and offers on social media during the COVID-19 pandemic.

\subsection{Advancing the area} Numerous future directions can be considered from this point forward. Our study assumes that requests and offers are geotagged. However, previous studies \cite{lamsal2021design,lamsal2023billioncov} have shown that less than 1\% of tweets are geotagged. To address this, our matching approach can be adjusted to utilize toponym extraction like in \cite{dutt2019utilizing,qazi2020geocov19}. However, toponym extraction introduces several challenges, including users being located at one place (Location A) while discussing another location (Location B), referred to as the \textit{Location A/B problem}, and the presence of multiple toponyms in the text. The initial problem can be addressed through binary classification, where text is analyzed to determine if the user mentions any origin location within the text \cite{lamsal2022did}. For example, in the sentence ``\textit{I love the weather in New York.}'' and ``\textit{I love the weather here in New York.}', the second sentence provides evidence that the user is in New York. As for the second issue, previous studies have utilized a majority vote method amongst toponyms to assign a location to tweets \cite{qazi2020geocov19}. However, challenges persist with this approach as it may still result in incorrect origin locations for the tweets.

Next, a multi-lingual dataset covering various crisis events is critical. This would enable the training of classifiers capable of understanding contextual clues associated with different crisis events, thereby enhancing the generalizability. The dataset in \cite{purohit2014emergency} is limited in that it only includes requests and offers within the context of a crisis. This leads to classifiers trained on this dataset identifying requests and offers that are outside of the scope of a crisis. As an example, texts such as ``Guys please help us spread the post, not the virus and TAG, REPOST \& SHARE'' are also getting classified as requests. While such texts are contextually requests, they are actually urging people to share information, not necessarily indicating a crisis situation. Furthermore, there is a need to refine the multi-lingual dataset to more accurately mirror the scripting style of social media texts. The current version lacks the use of abbreviations and informal language characteristics. The sentence encoder utilized in this study reaches its full potential when operating with informally constructed texts such as tweets.

Additionally, employing a unified embedding space for all languages simplifies the comparison of embeddings but leads to differences in performance of the encoder across languages, introducing biases in specific languages, primarily due to the scarcity of training data available for each language. Consequently, there exists an ongoing opportunity for enhancing the multi-lingual capacity of the sentence encoder.

\section*{Acknowledgements}
This research was undertaken using the LIEF HPC-GPGPU Facility hosted at the University of Melbourne, which was established with the assistance of LIEF Grant LE170100200. The cloud infrastructure required to maintain \textit{MegaGeoCOV Extended} over the last four years was provided by DigitalOcean.

\bibliographystyle{IEEEtran}
\bibliography{IEEEabrv,ref}

\end{document}